\documentclass[twoside,leqno,twocolumn]{article}

\usepackage[letterpaper]{geometry}
\usepackage{ltexpprt}

\usepackage{fancyhdr} 
\usepackage{times}
\usepackage{hyperref}
\usepackage{url}
\usepackage{graphicx}
\usepackage{amsmath,amssymb}
\usepackage{bm}
\usepackage{caption}
\usepackage{subcaption}
\usepackage{color}
\usepackage{mathtools}
\usepackage{tabularx}
\usepackage{booktabs, arydshln}
\usepackage[export]{adjustbox}

\begin{document}

\title{\Large DANR: Discrepancy-aware Network Regularization}


\author{Hongyuan You\thanks{Department of Computer Science, University of California, Santa Barbara
  (hyou@ucsb.edu, furkan@ucsb.edu, ambuj@ucsb.edu)}
\and Furkan Kocayusufoglu\footnotemark[1]
\and Ambuj K. Singh\footnotemark[1]}

\date{}

\maketitle

\begin{abstract} \small
\begin{quote}
Network regularization is an effective tool for incorporating structural prior knowledge to learn coherent models over networks, and has yielded provably accurate estimates in applications ranging from spatial economics to neuroimaging studies. 
Recently, there has been an increasing interest in extending network regularization to the spatio-temporal case to accommodate the evolution of networks. However, in both static and spatio-temporal cases, missing or corrupted edge weights can compromise the ability of network regularization to discover desired solutions. 
To address these gaps, we propose a novel approach---{\it discrepancy-aware network regularization} (DANR)---that is robust to inadequate regularizations and effectively captures model evolution and structural changes over spatio-temporal networks. 
We develop a distributed and scalable algorithm based on alternating direction method of multipliers (ADMM) to solve the proposed problem with guaranteed convergence to global optimum solutions. 
Experimental results on both synthetic and real-world networks demonstrate that our approach achieves improved performance on various tasks, and enables interpretation of model changes in evolving networks.
\end{quote}
\end{abstract}

\section{INTRODUCTION}


Network regularization is a fundamental approach to encode and incorporate general relationships among variables, which are represented as nodes and linked together by weighted edges that describe their local proximity. A significant amount of effort has been devoted to developing successful regularizations~\cite{tibshirani2005sparsity,sharpnack2012sparsistency,hallac2015network,wang2015trend} that take advantage of prior knowledge on network structures to enhance estimation performance for a variety of application settings, including image denoising~\cite{pang2017graph} and genomic data analysis~\cite{li2008network}.


We consider the general setting of network regularization, now proposed as a convex optimization problem defined on an undirected graph $\mathcal{G}(\mathcal{V}, \mathcal{E})$ with node set $\mathcal{V}$ and edge set $\mathcal{E}$:
\begin{align}
    \label{network-based-reg}
    \operatorname{min} \sum_{i\in\mathcal{V}} f_i(\bm{x}_i) + \lambda \sum_{(j,k)\in \mathcal{E}} \omega_{jk} \cdot g(\bm{x}_j,\bm{x}_k),
\end{align}
where on each node $i$, we intend to learn a local model $\bm{x}_i \in \mathbb{R}^{d}$, which in addition to minimizing a predefined {\it convex} loss function $f_i: \mathbb{R}^{d} \to \mathbb{R}$, carries certain relations to its neighbors, defined by the function $g: \mathbb{R}^{d} \times \mathbb{R}^{d} \to \mathbb{R}$.


In this paper, we focus on the particular setting where $g(\bm{x}_i,\bm{x}_j)$ = $\|\bm{x}_i-\bm{x}_j\|_2$, also known as a {\it sum-of-norms} (SON) regularizer. It assumes that the network is composed of multiple clusters, suggesting that all nodes within a cluster share the same consensus model ($\bm{x}_i = \cdots = \bm{x}_j$). As the observed data on each node may be sparse, the SON regularizer allows nodes to "borrow" observations from their neighbors to improve their own models, as well as to determine the network cluster to which they belong. The SON objective was first introduced in \cite{hocking2011clusterpath} 
for convex clustering problems and used off-the-shelf sequential convex solvers. Chi \& Lange \cite{chi2015splitting} adapt SON clustering to incorporate similarity weights with an arbitrary norm and solve the problem by both alternating direction method of multipliers (ADMM) or alternating minimization algorithm (AMA) in a parallel manner. More recently, Hallac et al. \cite{hallac2015network} point out that weighted SON (named as Network Lasso) allows for simultaneous clustering and optimization on graphs, and is highly suitable for a broad class of large-scale network problems.




In network regularization, a static weight assigned to an edge determines how strictly the difference between models on the corresponding nodes is being penalized, relative to the other node pairs in the network. 
However, unlike few scenarios in which network information is explicitly given, edge weights are usually unavailable or even infeasible to obtain in most real-world networks (e.g., gene regulatory networks \cite{liu2011controllability}).
Moreover, due to possible measurement errors and inaccurate prior knowledge, the assigned edge weights don't necessarily align with the underlying clustering structure of networks as shown in \cite{anselin2002under, harris2011search}. As a result, heavily relying on the edge weights in determining how much penalty should be applied to neighboring models may contaminate the discovered solutions.



Similar intuition applies when we extend to the temporal setting. Models on nodes of temporal networks usually change at the level of groups over time. However, some groups exhibit different evolution patterns than others \cite{kim2013nonparametric}. Moreover, the grouping structures themselves may evolve with time \cite{ho2011evolving}. As static regularization cannot capture such temporal evolution, a direct solution is to induce a time-varying local consensus by employing the SON objective on both spatial and temporal directions, but we face an more drastic problem: how to assign weights between snapshots.


Consequently, a framework that is robust to missing or corrupted edge weights in network-based regularization and adapts to spatio-temporal settings is desired. Note that without the assigned network weights, the setting in Eq. (\ref{network-based-reg}) degrades to simply applying isotropic regularization for all the edges. In this work, we propose a generic formulation, called the {\it discrepancy-aware network regularization} (DANR), which deploys a suitable amount of anisotropic network regularization in both spatial and temporal aspects. DANR infers models at each node per timestamp and can learn evolution of models and transitions of network structures over time. We develop an ADMM-based algorithm that adopts an efficient and distributed iterative scheme to solve problems formulated by the DANR, and show that the proposed solution obtains guaranteed convergence towards global optimal solutions. By applying to both synthetic and real-world datasets, we demonstrate the effectiveness of the proposed approach on various network problems.








\section{Problem Setting}

\subsection{Discrepancy-aware Network Regularization }


Since existing network-based regularizers rely on known weights on edges, the corresponding solutions get misled when the edge weights are erroneous or unknown. Directly learning the unknown edge weights by using the same regularizer as Eq. (\ref{network-based-reg}) would lead to an optimization problem:
\begin{small}
    \begin{align} \label{eq-failed-formulation}
    \underset{\bm{x},\bm{\omega}}{\operatorname{min~}}  \quad 
    \sum_{i \in \mathcal{V}} f_{i}(\bm{x}_{i}) + \lambda  \sum_{(j,k) \in \mathcal{E}} \omega_{jk} \left\|\bm{x}_{j}-\bm{x}_{k}\right\|_{2}
    \end{align}
\end{small}which yields the trivial all-zero solution of $\bm{\omega}$ and thus becomes unsatisfactory. Instead of imposing more sophisticated model-based or problem-dependent regularization as suggested in \cite{yu2016temporal}, we consider an alternative formulation that explicitly accounts for discrepancies between models on adjacent nodes:
\begin{small}
    \begin{multline}
     \label{eq-tvnl-tvpp2-s}
        \underset{\bm{x},\bm{\alpha}}{\operatorname{min~}} \quad \sum_{i \in \mathcal{V}} f_{i}(\bm{x}_{i}) + \lambda \cdot  \mathcal{R}_{S}(\bm{x},\bm{\alpha}) \\
        \mathcal{R}_{S}(\bm{x},\bm{\alpha}) = \mu \sum_{(j,k) \in \mathcal{E}} \omega_{jk}\left\|\bm{x}_{j}+\bm{\alpha}_{jk}-\bm{x}_{k}\right\|_{2} + (1-\mu) \|\bm{\alpha}\|_{1,p}
    \end{multline} 
\end{small}where we denote $\|\bm{\alpha}\|_{1,p} = \sum_{(j,k)\in \mathcal{E}} \|\bm{\alpha}_{jk}\|_p$ as the sum of $p$-norms of all $\bm{\alpha}_{jk}$'s. We set a penalty parameter $\lambda \geq 0$ to control the overall strength of network regularization, and a portion parameter $0 < \mu < 1$ to control the emphasis between the two terms in $\mathcal{R}_{S}(\bm{x},\bm{\alpha}) $. We name this formulation in Eq.~(\ref{eq-tvnl-tvpp2-s}) the {\it discrepancy-aware network regularization} (DANR).

\hspace{1em}In the first term of $\mathcal{R}_{S}$, we define a {\it discrepancy-buffering (DB) variable} $\bm{\alpha}_{jk} \in \mathbb{R}^d$ to denote the preserved discrepancy between local model parameters $\bm{x}_i$ and $\bm{x}_j$. More specifically, when an edge weight $\omega_{jk}$ is given but potentially imprecise or otherwise corrupted, $\bm{\alpha}_{jk}$ would compensate for abnormally large differences between models $\bm{x}_j$ and $\bm{x}_k$ to reduce the magnitude of $\omega_{jk}$-weighted edge penalty term in $\mathcal{R}_{S}$. The DB variable provides additional flexibility to its associated models, allowing them to stay adequately close to their own local solutions w.r.t. minimizing local loss functions, and avoid over-penalized consensus. When all $\omega_{jk}$'s are not given, $\bm{\alpha}_{jk}$'s enable solving Eq.~\ref{eq-tvnl-tvpp2-s} under anisotropic regularizations even with homogeneous weights.

The second term of $\mathcal{R}_{S}$ is the $\ell_{1,p}$-regularizer (with $1 < p < \infty$) \cite{vogt2012complete} that ensures sparsity at the group level. Note that we regard each variable vector $\bm{\alpha}_{jk}$ as a group ($\vert \mathcal{E} \vert$ groups in total). The first key intuition here is that the regularization on $\bm{\alpha}$ helps to exclude trivial solutions, that is $\bm{\alpha}_{jk} = \bm{x}_k - \bm{x}_j$. 
Second, it allows us to identify a succinct set of non-zero vectors $\bm{\alpha}_{jk}$'s, compensating for possible intrinsic discrepancies between two adjacent nodes.


To sum up, discrepancy-buffering variable $\bm{\alpha}_{jk}$'s are designed to elaborately adjust network regularization strength on all edges, and thus reduce negative effects of the unsquared norm regularizer. We will show in later sections that this modified formulation remains convex and  tractable via parallel optimization algorithms on large-scale networks.


\subsection{Distributed ADMM-based Solution}

In this section, we propose an ADMM-based algorithm for solving the ST-DANR problem in Eq.~(\ref{eq-tvnl-full}) and present the convergence and complexity of the proposed algorithm. The algorithm can be easily adapted to the spatial-only DANR problem in Eq.~(\ref{eq-tvnl-tvpp2-s}) by omitting temporal-related updates.


The ADMM method was originally derived in~\cite{gabay1976dual} and has been reformulated in many contexts including optimal control and image processing~\cite{peng2012rasl}. The method can be considered as combining augmented Lagrangian methods and the method of multipliers~\cite{boyd2011distributed}. 
It aims to solve optimization problems with two-block separable convex objectives in the following form of
{\small
    \begin{align} \label{admm-form}
        \operatorname{min}~ f(\bm{x}) + g(\bm{z}) \qquad ~s.t.~  A\bm{x}+B\bm{z}=\bm{c}
    \end{align}
}Observe that our proposed objective in Eq.~(\ref{eq-tvnl-full}) has a separable convex objective function: it can be reorganized into two-block separable convex objectives as in Eq.~(\ref{admm-form}), for which ADMM methods guarantee convergence to global optimal solutions~\cite{eckstein1992douglas}.

More precisely, to fit our problem into the ADMM framework, we first define {\small $\bm{x}$=$\lbrace \bm{x}_{j} \rbrace^{j \in V}$} for model parameter vectors on the given undirected network $\mathcal{G}$, and define consensus variables {\small $\bm{u}$ = $[\lbrace \bm{u}_{jk},\bm{u}_{kj} \rbrace^{(j,k)\in \mathcal{E}}]$} as copies of $\bm{x}$ in the spatial penalty term {\small $\mathcal{R}_{S}(\bm{x}, \bm{\alpha})$}. Then we rewrite Eq.~(\ref{eq-tvnl-full}) as follows: 
\vspace{-1ex}
\begin{small}
    \begin{align} \label{eq-tvnl-tvpp3-admm}
       \underset{\bm{x},\bm{u},\bm{\alpha}}{\operatorname{min}} \quad
        & \sum_{j\in \mathcal{V}} f_{j}(\bm{x}_{j}) + \lambda_1 (1-\mu_1) \|\bm{\alpha}\|_{1,p} \\
        & + \lambda_1 \mu_1 \sum_{(j,k)\in \mathcal{E}} \omega_{jk}\left\|\bm{u}_{jk}+\bm{\alpha}_{jk}-\bm{u}_{kj}\right\|_{2} \nonumber \\
        \text{~~s.t.~~} &
        \left[ \bm{u}_{jk}, \bm{u}_{kj} \right] ~~~= \left[ \bm{x}_{j}, \bm{x}_{k} \right], \quad (j,k) \in \mathcal{E} \nonumber
    \end{align}
\end{small}in which the first term corresponds to the $f$ block in Eq.~(\ref{admm-form}), and the remaining terms correspond to the $g$ block. The equality constraints on Eq.~(\ref{eq-tvnl-full}) are used to force consensus between variables $\bm{x}$ and $\bm{u}$. Next, we derive the augmented Lagrangian of Eq.~(\ref{eq-tvnl-tvpp3-admm}): 
\begin{small}
    \begin{align} 
    \label{eq-tvnl-tvpp3-lag}
        &\mathcal{L}_{\rho_1}(\bm{x},\bm{u},\bm{\alpha},\bm{\delta}) = \sum_{j\in V} f_{j}(\bm{x}_{j})
        + \lambda_1 (1-\mu_1) \|\bm{\alpha}\|_{1,p}
         \\
        &\hspace{3em} + \sum_{(j,k)\in E} \Big( \lambda_1 \mu_1 \omega_{j,k} \|\bm{u}_{jk}+\bm{\alpha}_{jk}-\bm{u}_{kj}\|_2
        \nonumber \\
        &\hspace{6em} +\frac{\rho_1}{2} \| \bm{x}_{j}-\bm{u}_{jk}+\bm{\delta}^{u}_{jk}\|_2^2-\frac{\rho_1}{2}\|\bm{\delta}^{u}_{jk}\|_2^{2}
        \nonumber \\
        &\hspace{6em} + \frac{\rho_1}{2}\|\bm{x}_{k}-\bm{u}_{kj}+\bm{\delta}^{u}_{kj}\|_2^2 - \frac{\rho_1}{2} \|\bm{\delta}^{u}_{kj}\|_2^2 \Big)
        \nonumber
    \end{align}
\end{small}where $\bm{\delta}$ = $[\lbrace \bm{\delta}^{u}_{jk},\bm{\delta}^{u}_{kj} \rbrace^{(j,k)\in \mathcal{E}}]$ are scaled dual variables for each equality constraint on the elements of $\bm{u}$. The parameters $\rho_1 > 0$ penalize the violation of equality constraints in the spatial and temporal domain~\cite{nishihara2015general} respectively. The iterative scheme of ADMM under the above setting can be written as follows, with $l$ denoting the iteration index:
{\small
    \begin{align}
        \label{eq-tvnl-tvpp3-admm-scheme}
        \begin{rcases}
            &\bm{x}^{(l+1)}=\underset{\bm{x}}{\operatorname{argmin~}}
            \mathcal{L}_{\rho_1}(\bm{x},(\bm{u},\bm{\alpha},\bm{\delta})^{(l)})  \\
            &(\bm{u},\bm{\alpha})^{(l+1)}=\underset{\bm{u},\bm{\alpha}}{\operatorname{argmin~}}
            \mathcal{L}_{\rho_1}(\bm{x}^{(l+1)},\bm{u},\bm{\alpha},\bm{\delta}^{(l)})  \\
            &\bm{\delta}^{(l+1)} = \bm{\delta}^{(l)} + (\tilde{\bm{x}}^{(l+1)}-\bm{u}^{(l+1)})
        \end{rcases}
    \end{align}
}where {\small $\tilde{\bm{x}}$=$[\lbrace \bm{x}_{j},\bm{x}_{k} \rbrace^{(j,k)\in \mathcal{E}}]$} is composed of replicated elements in $\bm{x}$, and thus has a one-to-one correspondence with elements in $\bm{u}$. 

Because the augmented Lagrangian in Eq.~(\ref{eq-tvnl-tvpp3-lag}) has a separable structure as well, we can further split the optimization above over each univariate element in $\bm{x}$ and $\bm{z}$. Next, we provide details for each ADMM update step.

\textbf{$\bm{x}$-Update.} In the first update step of our ADMM updating scheme, we can decompose the problem of minimizing $\bm{x}$ into separately minimizing $\bm{x}_{j}$ for each node $j$ :
\begin{small}
    \begin{align} 
    \label{eq-tvnl-tvpp3-admm-x-detail}
        \bm{x}_{j}^{(l+1)} =\underset{\bm{x}_{j}}{\operatorname{argmin~}} 
        \Big( f_{j}(\bm{x}_{j})+\frac{\rho_1}{2}\sum_{k \in N_{j}}  \|\bm{x}_{j} - \bm{u}_{jk}^{(l)} + \bm{\delta}_{jk}^{u(l)} \|_{2}^{2}  \Big)  \nonumber
    \end{align}
\end{small}
\vspace{-2ex}

\noindent As above equation suggests, $\bm{x}_{j}$ on node $j$ first receives local information from the corresponding consensus variables belonging to all of its neighbors $k \in N_{j}$, then updates its value to minimize the loss function and remain close to neighbouring consensus variables. Since all remaining regularizations are quadratic, the $\bm{x}$-update problem can be efficiently solved whenever the loss function $f_{j}$ has certain properties, such as strong convexity.

\textbf{($\bm{u},\bm{\alpha}$)-Update.} We can further decompose the ($\bm{z},\bm{\alpha}$)-update step in Eq.~(\ref{eq-tvnl-tvpp3-admm-scheme}) into subproblems on each edge (and its related variables $\bm{u}_{jk},  \bm{u}_{kj}, \bm{\alpha}_{jk}$) as follows, which can be solved in parallel:
{\small
    \label{eq-tvnl-tvpp3-admm-u-alpha-detail}
    \begin{multline}
        (\bm{u}_{jk}^{(l+1)},  \bm{u}_{kj}^{(l+1)}, \bm{\alpha}_{jk}^{(l+1)}) = \\
        \operatorname{argmin~} \Big( \lambda_1 (1-\mu_1) \| \bm{\alpha}_{jk} \|_p + \lambda_1 \mu_1 \omega_{jk}\|\bm{u}_{jk}+\bm{\alpha}_{jk}-\bm{u}_{kj}\|_{2}  \\
        \hspace{1em} + \frac{\rho_1}{2}\|\bm{x}_{j}^{(l+1)}-\bm{u}_{jk} + \bm{\delta}^{u(l)}_{jk} \|_{2}^{2}  + \frac{\rho_1}{2} \|\bm{x}_{k}^{(l+1)}-\bm{u}_{kj} + \bm{\delta}^{u(l)}_{kj} \|_{2}^{2} \Big)
    \end{multline}
}Notice that the sum of convex functions which are defined on different sets of variables preserves the convexity. To minimize the objective with $\bm{u}_{jk},\bm{u}_{kj}$ and $\bm{\alpha}_{jk}$ simultaneously, the convexity of Eq.~(\ref{eq-tvnl-tvpp3-admm-u-alpha-detail}) motivates us to adopt an alternating descent algorithm, which minimizes each component iteratively with respect to $(\bm{u}_{jk},\bm{u}_{kj})$ and $\bm{\alpha}_{jk}$ while holding the other fixed. In detail, we solve Eq.~(\ref{eq-tvnl-tvpp3-admm-adm-u}) and Eq.~(\ref{eq-tvnl-tvpp3-admm-adm-alpha}) iteratively until convergence is achieved:
\vspace{-0.1in}
\begin{small} 
    \begin{align}
        \label{eq-tvnl-tvpp3-admm-adm-u}
        & (\bm{u}_{jk}^{(l'+1)}, \bm{u}_{kj}^{(l'+1)}) = {\operatorname{argmin}} \Big(  \lambda_1 \mu_1 \omega_{jk}\|\bm{u}_{jk} +\bm{\alpha}_{jk}^{(l')}-\bm{u}_{kj}\|_{2} \\
        & +\frac{\rho_1}{2} \|\bm{x}_{j}^{(l+1)}-\bm{u}_{jk} + \bm{\delta}^{u(l)}_{jk}\|_{2}^{2} +  \frac{\rho_1}{2}\|\bm{x}_{k}^{(l+1)}-\bm{u}_{kj} + \bm{\delta}^{u(l)}_{kj} \|_{2}^{2} \Big) \nonumber
    \end{align}
\end{small}
\vspace{-3ex}
\begin{small}
    \begin{align}
        \label{eq-tvnl-tvpp3-admm-adm-alpha}
            \bm{\alpha}_{jk}^{(l'+1)} &= {\operatorname{{\footnotesize argmin}}} \Big(
            (1-\mu_1) \| \bm{\alpha}_{jk} \|_p \\
            & \hspace{2em}+ \mu_1 \omega_{jk}\|\bm{u}_{jk}^{(l'+1)}+\bm{\alpha}_{jk}-\bm{u}_{kj}^{(l'+1)}\|_{2}   \Big) \nonumber
    \end{align}
\end{small}
\vspace{-0.15in}

\noindent where $l'$ denoting the iteration index of alternating descent. Further, we can utilize the analytic solution to speed up the calculation of $\bm{u}_{jk}$ and $\bm{u}_{kj}$:
\begin{lemma} Problem (\ref{eq-tvnl-tvpp3-admm-adm-u}) has a closed-form solution:
    \begin{small}
        \begin{equation}
            \begin{rcases}
                \bm{u}_{jk}^{(l'+1)} = (1-\theta)\bm{a}+\theta \bm{b} - \theta \bm{\alpha}_{jk}^{(l')} \qquad \nonumber \\
                \bm{u}_{kj}^{(l'+1)} = \theta \bm{a} + (1-\theta) \bm{b} + \theta \bm{\alpha}_{jk}^{(l')} \qquad \nonumber
            \end{rcases}
        \end{equation}
    \end{small}where we denote {\small $ \bm{a} = \bm{x}_{j}^{(l'+1)}+{\bm{\delta}_{jk}^{u}}^{(l')}$}, {\small $\bm{b} = \bm{x}_{k}^{(l'+1)}+{\bm{\delta}_{kj}^{u}}^{(l')}$}, {\small $c = \lambda_1 (1-\mu_1)\omega_{jk}$}, and {\small $\theta = c/( \rho_1 \| \bm{a}-\bm{b} + \bm{\alpha}_{jk}^{(l')}\|_2)$} for simplification. Details in Appendix B.
\end{lemma}

\textbf{$\bm{\delta}$-Update.} In the last step of our ADMM updating scheme, we have fully independent update rules for each scaled dual variable {\small $\bm{\delta}_{jk}^u$} as follows:
{\small
    \begin{align}
        \label{eq-tvnl-tvpp2u-acsx-admm-steps-delta}
            {\bm{\delta}_{jk}^u}^{(l+1)} &= {\bm{\delta}_{jk}^u}^{(l)} 
                    + ( \bm{x}_{j}^{(l+1)}-\bm{u}_{jk}^{(l+1)}) 
    \end{align}
}Finally, we present the pseudo-code of our DANR approach in Appendix D.


\textbf{Stopping Criterion and Global Convergence.} For our ADMM iterative scheme in Eq.~(\ref{eq-tvnl-tvpp3-admm-scheme}), we use the norm of primal residual $\bm{r}^{(l)} = \tilde{\bm{x}}^{(l)} - \bm{u}^{(l)}$ and dual residual $\bm{s}^{(l)} = \rho_1(\bm{u}^{(l)}-\bm{u}^{(l+1)})$ as the termination measure. The optimality condition~\cite{boyd2011distributed} of ADMM shows that if both residuals are small then the objective suboptimality must be small, and thus suggests $\|\bm{r}^{(l)}\|_2 \leq \epsilon_1 \wedge \|\bm{s}^{(l)}\|_2 \leq \epsilon_2$ as a reasonable stopping criterion. Convex subproblems in $\bm{x}$-update and $(\bm{u},\bm{\alpha})$-update need iterative methods to solve as well. The stopping criterion for these subproblems is naturally to keep iteration differences below thresholds, {\it i.e.} in $(\bm{u},\bm{\alpha})$-update, we require ${\small \|\Delta \bm{u}^{(l')}\|_2 \leq \epsilon'}$ and ${\small \|\Delta \bm{\alpha}^{(l')}\|_2 \leq \epsilon'}$. 

\begin{lemma}
Our ADMM approach to solve the DANR problem is guaranteed to converge to the global optimum. Details in Appendix C.
\end{lemma}

\textbf{Computational Complexity.} Let $N_c$ denote the number of iterations that ADMM takes to achieve an approximate solution $\hat{\bm{x}}$ with an accuracy of $\epsilon_x$. Based on the convergence analysis in \cite{he2015non}, the time complexity scales as $O(1/\epsilon_x)$, which in our problem mostly depends on the properties of cost functions $f_{jt}$'s. Assume that all convex subproblems in $x$-update and $(u,\alpha)$-update are solved by general first-order gradient descent methods that take $N_x$ and $N_\alpha$ iterations to converge respectively, the overall complexity of the algorithm is therefore $O \big( N_c \big( N_x  \vert \mathcal{V} \vert +  N_\alpha \vert \mathcal{E} \vert + \vert \mathcal{E} \vert \big) \big)$.
\section{Extension to Spatio-temporal Setting }

The aforementioned shortcomings with pre-defined edge weights accentuate when we consider temporal networks since acquiring explicit temporal edge weights is usually not feasible. Thus, discrepancy-buffering variables are also crucial for the temporal setting. 

\begin{figure}[t]
    \centering
    \begin{subfigure}[]{0.5\textwidth}
    \centering
        \includegraphics[width=\textwidth, clip=true]{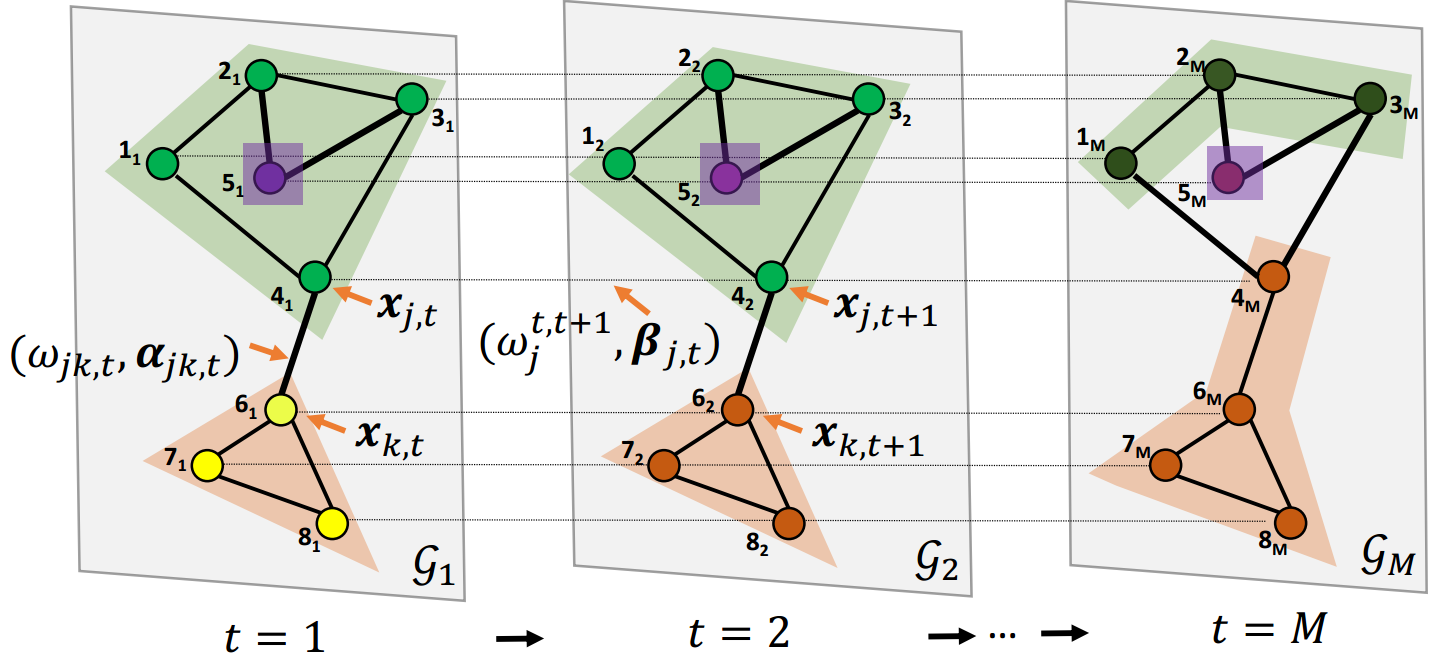}
    \end{subfigure}
    \vspace{-2ex}
    \caption{\small Overview of our problem setting in temporal networks. \vspace{-2ex}}
    \label{fig:tvnl_diag}
    \vspace{-0.2in}
\end{figure}

Consider a temporal undirected network $\mathcal{G}$ consisting of $M$ sequential network snapshots $\lbrace \mathcal{G}_1, \mathcal{G}_2, \ldots, \mathcal{G}_M\rbrace$, where each network snapshot $\mathcal{G}_t$ contains a node set $\mathcal{V}_t$ and an edge set $\mathcal{E}_t$. We denote $\omega_{j,k}^t$ as the weight of spatial edge $(j_t,k_t)$ between nodes $j_t$ and $k_t$ within snapshot $\mathcal{G}_t$, and $\omega_{j}^{t,t+1}$ for the weight of temporal edge $(j_t,j_{t+1})$ that links node $j_t$ with $j_{t+1}$ across snapshots $\mathcal{G}_t$ and $\mathcal{G}_{t+1}$ (shown in Fig.\ref{fig:tvnl_diag}). On each node $j_t$, a {\it convex} loss function $f_{j,t}: \mathbb{R}^{d} \to \mathbb{R}$ is given to measure the fitness of a local model parameterized by $\bm{x}_{j,t} \in \mathbb{R}^{d}$. With sparse observations for all snapshots, a straightforward task is to find jointly optimal models for every node and every timestamp under the regularization for both network topology and temporal evolution. We can propose an extension of the DANR formulation to spatio-temporal networks, referred as ST-DANR: \vspace{-1ex}
{\small
    \begin{equation}
        \label{eq-tvnl-full}
        \underset{\bm{x},\bm{\alpha},\bm{\beta}}{\operatorname{min}}
        \quad \sum_{j \in \mathcal{V}} \sum_{t=1}^{M} f_{j,t}(\bm{x}_{j,t}) +  \lambda_1 \cdot \mathcal{R}_{S}(\bm{x}, \bm{\alpha})  + \lambda_2 \cdot \mathcal{R}_{T}(\bm{x}, \bm{\beta}) 
    \end{equation}
}
\vspace{-2ex}

\noindent where we define the discrepancy-aware regularizers as:
\vspace{-1ex}
{\small 
\begin{align}
\vspace{-4ex}
\label{eq-tvnl-sreg}
\mathcal{R}_{S}(\bm{x}, \bm{\alpha}) 
& = \mu_1 \sum_{t = 1}^{M} \sum_{(j_t,k_t) \in \mathcal{E}_t} \left( \omega_{jk}^{t}\left\|\bm{x}_{j,t}+\bm{\alpha}_{jk,t}-\bm{x}_{k,t}\right\|_{2} \right) \nonumber \\
& + (1-\mu_1) \|\bm{\alpha}\|_{1,p} \qquad \text{(spatial penalty term)}  \\
\label{eq-tvnl-treg}
\mathcal{R}_{T}(\bm{x}, \bm{\beta}) 
& = \mu_2 \sum_{j \in \mathcal{V}} \sum_{t = 1}^{M-1} \left(  \omega_{j}^{t,t+1} \| \bm{x}_{j,t} + \bm{\beta}_{j,t} - \bm{x}_{j,t+1} \|_2 \right) \nonumber \\
& + (1-\mu_2) \|\bm{\beta}\|_{1,p} \quad \text{(temporal penalty term)}
\end{align} 
}
\vspace{-4ex}

\noindent As we are interested in the heterogeneous evolution of nodal models across time, the chosen unsquared norms in spatial and temporal regularization terms $\mathcal{R}_{S}(\bm{x}, \bm{\alpha}) $ and $\mathcal{R}_{T}(\bm{x}, \bm{\beta}) $ enforce piecewise consensus in both spatial and temporal aspects, which indicates abrupt changes of regional models or sudden transitions in network structures at particular timestamps, and also implies valid persistent models in the remaining segments of time \cite{hallac2017network}. Beside the spatial DB variables $\bm{\alpha}_{jk,t}$ in $\mathcal{R}_{S}(\bm{x}, \bm{\alpha}) $, we define another set of temporal DB variables $\bm{\beta}_{j,t} \in \mathbb{R}^d$ in $\mathcal{R}_{T}(\bm{x}, \bm{\beta}) $ to compensate for inadequate regularizations caused by temporal fluctuations. For example, when modeling housing prices in a city, $\bm{\beta}_{j,t}$ would tolerate anomalous short-term fluctuations in prices, while large values of $\bm{\alpha}_{jk,t}$ might be found near boundaries of disparate neighborhoods.

\textbf{Adaptation to Streaming Data.} In many applications, observed network snapshots arrive in a streaming fashion. Instead of recomputing all models on past snapshots whenever a new snapshot is observed, incorporating existing models to facilitate the new incoming snapshot is more efficient. Assume that we have learned models $\lbrace \hat{\bm{x}}_{j,t} \rbrace_{j \in \mathcal{V}}$ for the $\tau$-th snapshot, we then read observations of the next $m$ snapshots indexed by $\tau+1, \cdots, \tau+m$ and attempt to learn models for them. To this intent, we deploy our ST-DANR formulation in Eq.~(\ref{eq-tvnl-full}) on a limited time interval $t = \tau, \cdots,\tau+m$, and then integrate established model at $t=\tau$ by fixing $\lbrace \bm{x}_{j,\tau} \rbrace_{j \in \mathcal{V}}$ = $\lbrace  \hat{\bm{x}}_{j,\tau} \rbrace_{j \in \mathcal{V}}$ and solving the remaining variables $\lbrace \bm{x}_{j,t} \rbrace_{j \in \mathcal{V}}^{t=\tau+1,\cdots, \tau+m}$ in the corresponding problem. Note that in our setting, temporal messages are only transferable through consecutive snapshots, meaning that fixing the $\tau$-th snapshot is the same as fixing all past snapshots. 

Lastly, we point out that distributed ADMM-based solution for DANR (\S 2.2) can be adapted to ST-DANR, for either batch formulation (Eq.~(\ref{eq-tvnl-full})) or streaming-case formulation, which readers can refer to in Appendix E.



\textbf{Temporal Evolutionary Patterns.}
In this work, we consider the unsquared edge penalty term $\| \bm{x}_{j,t} + \bm{\beta}_{j,t} - \bm{x}_{j,t+1} \|_2$ in $\mathcal{R}_{T}$ since it enforces temporal reconstruction, indicating sharp transitions or change points at particular timestamps. Similar to the spatial case, there could be multiple available alternatives to enforce different types of evolutionary patterns \cite{hallac2017network}. For example, replacing the unsquared norm with squared norm form $\| \bm{x}_{j,t} + \bm{\beta}_{j,t} - \bm{x}_{j,t+1} \|_2^2$ will yield smoothly varying models along consecutive snapshots, and has been well studied in \cite{wang2014low, dang2017subnetwork}.
\section{EXPERIMENTS}

In this section, we present an experimental analysis of the proposed method (DANR) and evaluate its performance on classification and regression tasks. First, we employ a synthetic dataset (\S\ref{dataset_synthetic}) and two housing price datasets (BAY and SAC in \S\ref{dataset_housing}) to demonstrate the set of scenarios and applications that are particularly good fit for DANR. 
In addition to an improvement on the estimation task, DANR learns fine-grained neighborhood boundaries and reveals interesting insights into the network's heterogeneous structure.
Lastly, in order to validate the efficacy of ST-DANR in the temporal setting, we conduct experiments with the geospatial and temporal database of Northeastern US lakes (\S\ref{dataset_lake}), from which we aim to estimate the water quality of lakes over 10 years. 

\textbf{Baseline Algorithms.} In spatial network scenarios (\S 4.1 \& \S 4.2), we compare DANR against three baselines: network lasso (NL) \cite{hallac2015network}, robust multi-task feature learning (rMTFL) \cite{gong2012robust}, factorized multi-task learning (FORMULA) \cite{xu2015formula}. In spatial-temporal network scenarios (\S 4.3), we compare ST-DANR with two widely-applied temporal regularizers in the literature, which are detailed in \S 4.3.

\textbf{Parameters Setting. } For both NL and DANR, we tune $\lambda$ parameters from $10^{-3}$ to $10^2$ where $\lambda^{n+1} = 1.3 \lambda^{n}$. Concerning the DANR, for each value of $~\lambda$, we further tune $\mu$ parameters from $0.3$ to $1$, where  $\mu^{n+1} = \mu^{n} + 0.02$. Lastly, we set $p=3$. We follow the same strategy for both spatial (DANR) and spatio-temporal (ST-DANR) variants of the proposed method. For all penalty parameters in rMTFL and FORMULA, we tune them in the same way as $\lambda$ in DANR. In addition, we vary the number of factorized base models in FORMULA from $1$ to $50$ to achieve its best performance.  For each dataset, we standardize all features and response variables to zero mean and unit variance. 

\vspace{-0.05in}
\subsection{Node Classification with Limited Data} \label{dataset_synthetic}

Following previous work~\cite{hallac2015network}, we first experiment with a synthetic network in which each node attempts to solve its own support vector machine (SVM) classifier, but only has a limited amount of data to learn from. We further split the nodes into clusters where each cluster has an underlying model, i.e., nodes in the same cluster share the same underlying model. Our aim is to learn accurate models for each node by leveraging their network connections and limited observations. Note that the neighbors with different underlying models provide misleading information to each other, which potentially harms the overall performance. Therefore, we later investigate the robustness of the DANR with a varying ratio of such malicious edges.

\textbf{Synthetic Network Generation. } 
We create a network of 100 nodes, split into 5 ground-truth communities $C^1$, $C^2$, $\cdots$, $C^5$. Each of these communities consists of 20 nodes. We denote the community of a node $i$ with $C_i$. Next, we form the network connections by adding edges between nodes based on the following criterion: The probability of adding an edge between any node pair $(i,j)$ is 0.5 if $C_i = C_j$, i.e., the edge connects nodes from the same community (\emph{intra-community edge}). Otherwise, we set the probability as 0.02 if $C_i \neq C_j$, i.e., the edge connects nodes from different communities (\emph{inter-community edge}). The resulting synthetic network $G_0 (\mathcal{V}, \mathcal{E})$ has 100 nodes and 574 edges, with 82\% of them being intra-community edges. Each ground-truth community $C^k$ has an underlying classifier model $\bm{X}^{k} \in R^{10}$, drawn independently from a normal distribution of zero mean and unit variance, where $k \in [1, \cdots, 5]$. Next, we generate 5 random training example pairs $(\bm{w},y)$ per node,  where $\bm{w} \in R^{10}$ denotes an observation and $y \in \{-1,1\}$ denotes the ground-truth value to estimate. The ground-truth value $y$ for each observation $\bm{w}$ is then computed using the underlying classifier of the community that a node belongs to:
\vspace{-0.1in}
\begin{align}
    y = \text{sign} ( \bm{w}^{T} \bm{X}^{k} + \eta ) 
    \label{synthetic_y}
\end{align}
\vspace{-4ex}
 
\noindent where $\bm{X}^{k} \in R^{10}$ represents the corresponding weight of the classifier w.r.t. observations, while $\eta$ is the noise. All observations and noises are drawn independently from a normal distribution for each data point. Note that 5 observations are insufficient to accurately estimate a model $\bm{x}_i \in R^{10}$. Hence for this setting, the network connections play an important role as nodes can ``borrow'' training examples from its neighbors to improve their own models.

\textbf{Optimization Problem.}
The objective function $f_i$ on each node is formulated by the soft margin SVM objective, where node $i$ estimates its model (i.e. separating hyperplane) $\bm{x}_i \in R^{10}$ using its five training examples $(\bm{w},y)$ as follows:
\vspace{-2ex}
{\small
\begin{align}
    \nonumber
    \min \Big(\frac{1}{2} \bm{x}_i^T \bm{x}_i + C \sum_{l=1}^{5} \xi_{l} \Big)
    \quad  \text{s.t.  } y^{(l)} (\bm{x}_i^T \bm{w}^{(l)}) \geqslant (1-\xi_{l}), ~\xi_{l} \geqslant 0
\end{align} 
}
\vspace{-3ex}

\noindent where $\xi_{l}$'s are slack variables that accounts for non-separable data. The constant $C$ controls the trade-off between minimizing the training error and maximizing the margin. We set $C$ to 0.75, which we empirically find to perform well.

\textbf{Test Results.} In following subsections, we first thoroughly inspect DANR and its prototype method NL on synthetic network $G_0$, in order to show the benefits of introducing discrepancy-buffering variable $\bm{\alpha}$. Later in \S 4.1.2, we evaluate the performance and robustness of DANR and all three baseline methods under more noisy scenarios. To assess the performance, prediction accuracies are computed over a separate set of 1000 test pairs (10 per node). The test pairs are again randomly generated using Eq.~(\ref{synthetic_y}).
\vspace{-0.15in}

\subsubsection{Effect of Discrepancy-Buffering Variables.} \label{compare_with_nl}

\begin{figure}[ht]
    \centering
    \begin{subfigure}[b]{0.3\textwidth}
    \centering
        \includegraphics[width=\textwidth, clip=true]{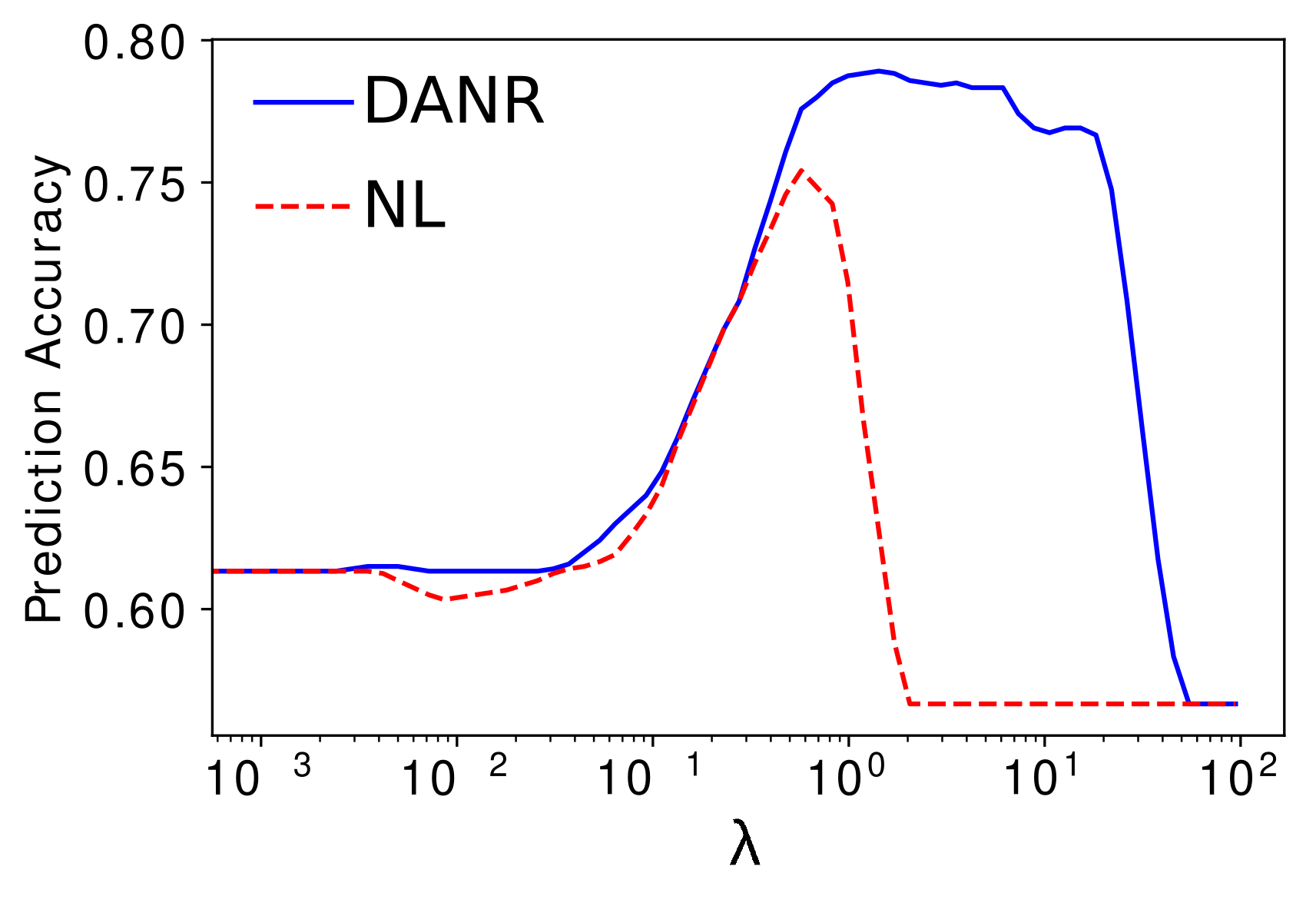}
    \end{subfigure}
    \vspace{-0.15in}
    \caption{\small Prediction accuracy comparison with varying $\lambda$ values.}
    \label{fig:lambda_comparison}
    \vspace{-0.15in}
\end{figure}

We compare DANR with NL and report the prediction accuracy with varying $\lambda$ values in Figure~\ref{fig:lambda_comparison}. Recall that the proposed method introduces the $\mu$ parameter which is coupled with the $\lambda$ parameter. Therefore in order to have an adequate comparison between DANR and NL, we modify our $\lambda$ parameter to be $\tilde{\lambda} = \lambda / \mu$. Doing so ensures that both methods apply the same amount of penalty to their network regularization terms (See Eq.~\ref{eq-failed-formulation} and \ref{eq-tvnl-tvpp2-s}). For each value of $~\lambda$, we vary $\mu$ from 0.3 to 1 and report the highest accuracy achieved.

\begin{figure}[h]
    \vspace{-0.05in}
    \begin{subfigure}[b]{0.23\textwidth}
        \includegraphics[width=\textwidth, clip=true]{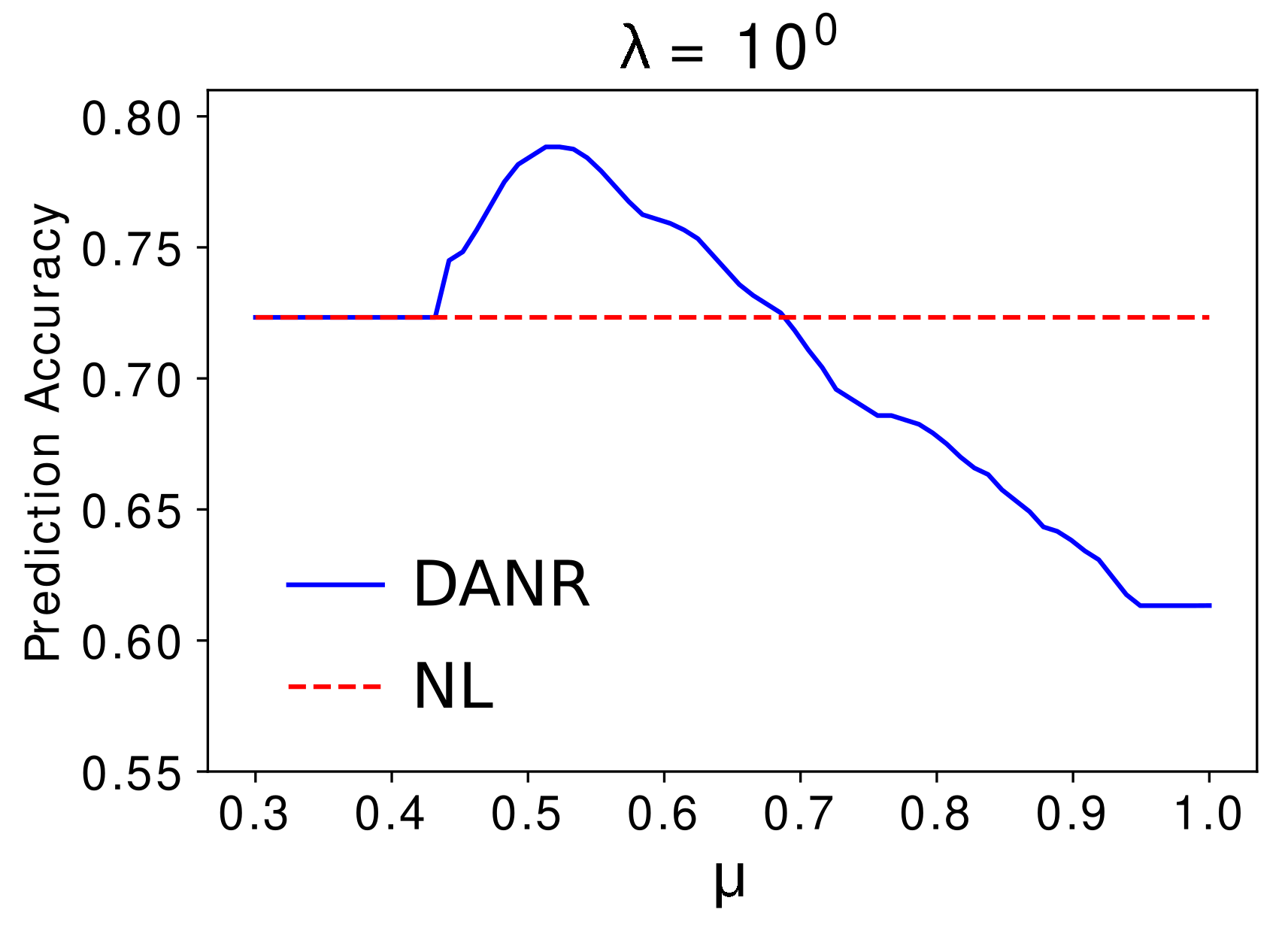}
    \end{subfigure}
    \begin{subfigure}[b]{0.23\textwidth}
        \includegraphics[width=\textwidth, clip=true]{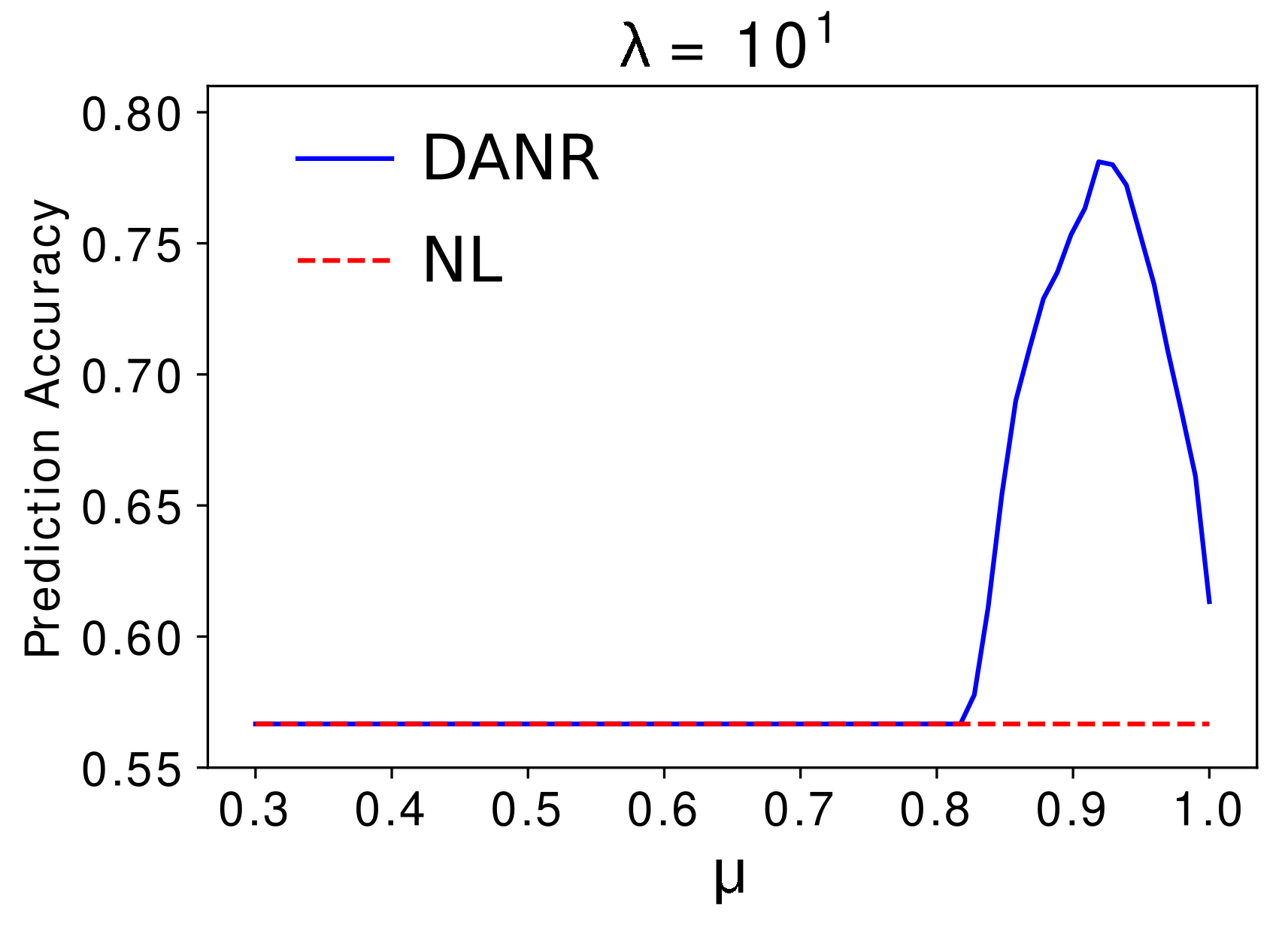}
    \end{subfigure}
    \vspace{-0.15in}
    \caption{\small Effect of $\mu$ on the prediction accuracy, with a fixed $\lambda$.}
    \label{fig:mu_comparison}
    \vspace{-0.2in}
\end{figure}

We first briefly mention the desired behaviors of the NL method and later accentuate its drawbacks. As shown by Figure~\ref{fig:lambda_comparison}, for small $\lambda$ values, the problem reduces to solving the local optimization problem where each node estimates its model solely based on its limited observations. This achieves 61.3\% accuracy on the test set. Furthermore, as $\lambda$ increases, the NL penalty enforces nodes to form clusters, where nodes in the same cluster share the same model. This behavior firmly improves the test performance with prediction accuracy rising to 75.3\%. Right after the peak performance is reached however, a slight increase in $\lambda$ causes a sudden drop in the performance and the problem reduces to solving the global optimization problem over the entire network. Therefore when $\lambda > \lambda_{critical}$, the problem converges to a common model for all the nodes in the network , i.e., $\bm{x}_i = \bm{x}_j$ for $\forall i, j \in \mathcal{V}$. This further drops the prediction accuracy to 56.4\% on the test set. 

The observed rapid drop in performance implies that the NL method is highly sensitive to the $\lambda$ parameter. There is only a narrow window of $\lambda$ values that result in a proper clustering of the network. As a result, one needs to excessively tune the $\lambda$ parameter around $\lambda_{critical}$ to find its optimal value. In addition, such clustering performance is also highly affected by the volume of ``malicious'' (inter-cluster) edges in the network, as shown in \S\ref{malicious_edges}. 

From Figure~\ref{fig:lambda_comparison}, we can observe that the DANR approach exhibits improved performance thanks to its discrepancy-buffering variables, $\alpha$, which allow our regularization term to better exploit the good edges while providing additional tolerance towards bad edges. As a result, DANR achieves 79.1\% accuracy on the test set, outperforming the best baseline method by 3.8\%. Another important observation is that DANR provides a much wider window for selecting near-optimal $\lambda$ than the baseline approach. That being said, we now analyze how the \emph{$\mu$} parameter couples with the $\lambda$ parameter, and further present its effect on DANR's performance. Figure~\ref{fig:mu_comparison} displays the prediction accuracy vs. $\mu$, with a fixed $\lambda = 10^{0}$ and $10^{1}$. Intuitively, as $\lambda$ increases, the optimal value of $\mu$ also increases. The reason behind is that as $\lambda$ increases, the more non-zero discrepancy-buffering variables (parameterized by ($1-\mu$)) are needed to persist the accurate clustering formation since high values of $\lambda$ forces global consensus over the network.





\begin{figure}[h]
    \vspace{-2ex}
    \centering
    \begin{subfigure}[b]{0.36\textwidth}
    \centering
        \includegraphics[width=\textwidth, clip=true]{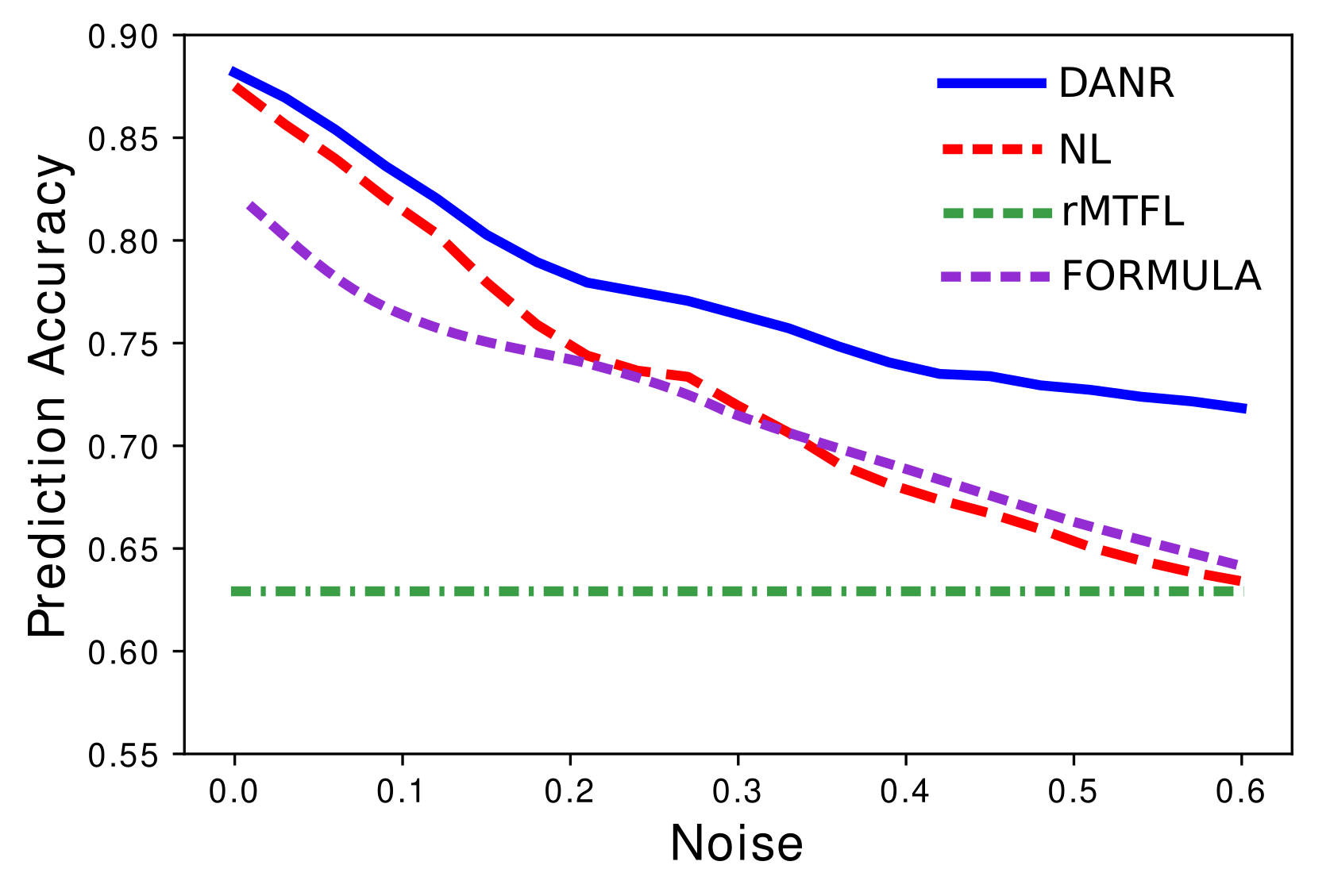}
    \end{subfigure}
    \vspace{-4ex}
    \caption{\small Prediction accuracy comparison with varying noise.\vspace{-3ex}}
    \label{fig:noise_comparison}
\end{figure}
\vspace{-2ex}
\subsubsection{Robustness to Malicious Edges.} \label{malicious_edges}
Note that the ratio of inter-community edges in the earlier synthetic network $G_0$ is 18\%. Here we use term \emph{noise} for the ratio of such malicious edges in the network. Taking into account that the amount of noise is critical in learning accurate models for nodes in a network, we now investigate the efficacy of DANR w.r.t. varying noise. First, we remove all inter-community edges and later iteratively add malicious edges to the network. Then we solve the same problem with the noise varying from 0 to 0.6, and report prediction accuracies of all methods in Figure~\ref{fig:noise_comparison}.
As shown, the performance of DANR, NL and FORMULA drops in the absence of noise. Meanwhile the pure multitask method rMTFL maintains the same low accuracy, since it doesn't utilize the network information. However, as we increase the noise, DANR starts to outperform NL and FORMULA, where the gap between the models' performances becomes larger at a higher ratio of malicious edges. This confirms that compared to the baseline methods, the DANR exhibits more robust performance in noisy settings thanks to its discrepancy-buffering variables.

\vspace{-0.1in}
\subsection{Spatial: Housing Rental Price Estimation} \label{dataset_housing}

In this section, our goal is to jointly (i) estimate the rental prices of houses by leveraging their geological location and the set of features; (ii) discover boundaries between neighborhoods. The intuition is that the houses in the same neighborhood often tend to have similar pricing models. However, such neighborhoods can have complex underlying structures (as described later in this section), which imposes additional challenges in learning accurate models.

\textbf{Dataset.} We experiment with two largely populated areas in Northern California: the Greater Sacramento Area (SAC) and the Bay Area (BAY). The anonymized data is provided by the property management software company {\it Appfolio Inc}, from which we further sample houses with at least one signed rental agreement during the year 2017. The resulting dataset covers 3849 houses in SAC and 1498 houses in BAY. Concerning the houses that have more than one rental agreement signed during 2017, we average the rental prices listed in all the agreements. Each house holds the information about its location (latitude/longitude), number of bedrooms, number of bathrooms, square footage, and the rental price. We regard these areas (SAC and BAY) as two separate datasets for the remainder of this section. We also randomly split 20 \% of the houses in each dataset for testing.

\textbf{Network Construction.} After excluding test houses from both datasets, we construct two networks (one for each dataset) based on the houses' locations. An undirected edge exists between node $i$ and $j$, if at least one of them is in the set of 10 nearest neighbors of the other. (Note that node $j$ being one of the 10 nearest neighbors of $i$ doesn't necessarily imply that node $i$ is also in the set of nearest neighbors of $j$.) In the resulting networks, the average degree of a node is 12.16 for SAC and 12.04 for BAY. Additionally, we construct two versions of these networks, \emph{weighted} and \emph{unweighted}. While the weighted network has edge weights inversely proportional to the distances between houses, the unweighted network ignores the proximity between houses, and thus weights on all the edges are 1.

\textbf{Optimization Problem.} The model at each house estimates its rental price by solving a linear regression problem. More specifically, at each node $i$ we learn a 4-dimensional model $\bm{x}_i = [a_i,b_i,c_i,d_i]$. $\bm{x}$ simply represents the coefficients of each feature, which later is used to estimate the rental price $p_i$ as follows:
\vspace{-0.1in}
{\small
\begin{align}
    \nonumber
    \overline{p_i} = a_i \cdot (\# bedrooms) + b_i \cdot (\# bathrooms) \\
    \nonumber
    + c_i \cdot (square footage) + d_i,
\end{align}
}where $d_i$ is the bias term. The training objective is
{\small
\begin{align} 
    \min_{\bm{x,\alpha}}~ \sum_{i \in \mathcal{V}} \|\overline{p_i} - p_{i}\|_2^2 + c \cdot \|\bm{x}_i\|_2^2 + \lambda \cdot \mathcal{R}_{S}(\bm{x}, \bm{\alpha}) \nonumber
\end{align}
}where $\mathcal{R}_S$ represents the proposed network regularization term (see Eq.~\ref{eq-tvnl-tvpp2-s}), while $c$ is the regularization weight to prevent over-fitting.

\begin{figure}[b]
    \vspace{-3ex}
    \centering
    \begin{subfigure}[h]{0.232\textwidth}
    \centering
        \includegraphics[width=\textwidth, clip=true]{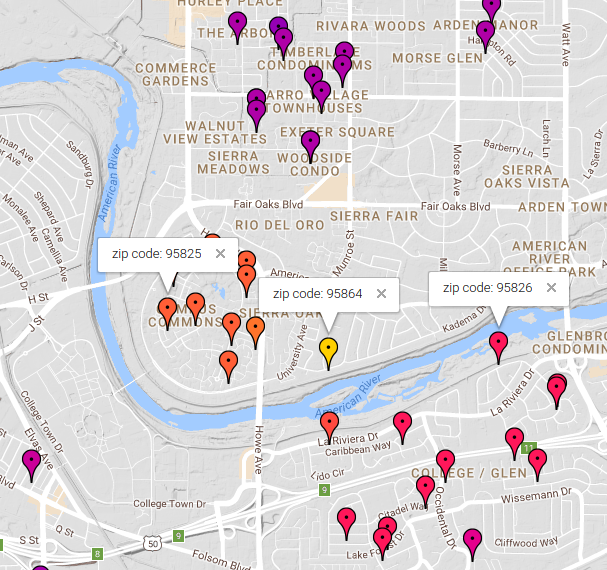}
    \end{subfigure}
    \begin{subfigure}[h]{0.23\textwidth}
    \centering
        \includegraphics[width=\textwidth, clip=true]{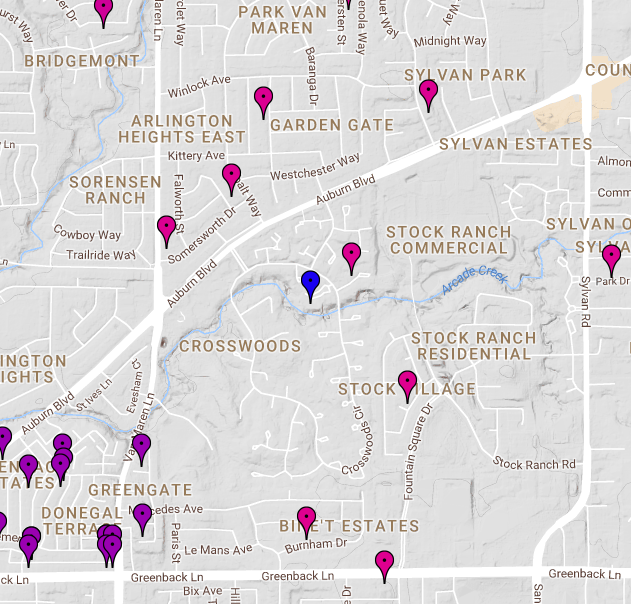}
    \end{subfigure}
    \vspace{-0.1in}
    \caption{\small Examples of complex neighborhood structures captured by the DANR. In left, the house shown in yellow resides at the border of three different area codes. In right, the creek side house (colored in blue) differs from all of its neighbors, possibly due to its appealing location.} \vspace{-8ex}
    \label{fig:sac_zoom1}
\end{figure}

\begin{figure*}[t]
    \centering
    \begin{subfigure}[h]{0.33\textwidth}
    \centering
        \includegraphics[width=\textwidth, clip=true]{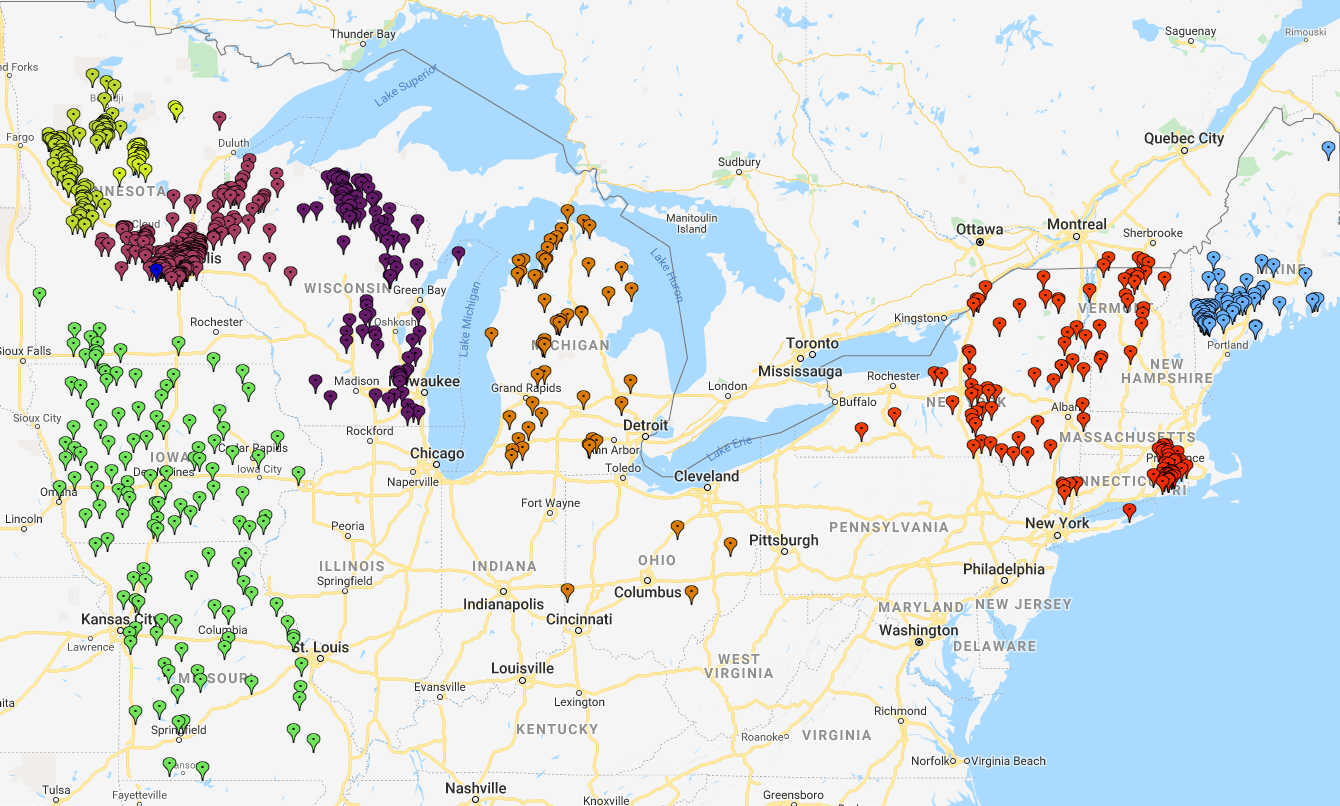}
        \caption{2000}
    \end{subfigure}
    \begin{subfigure}[h]{0.33\textwidth}
    \centering
        \includegraphics[width=\textwidth, clip=true]{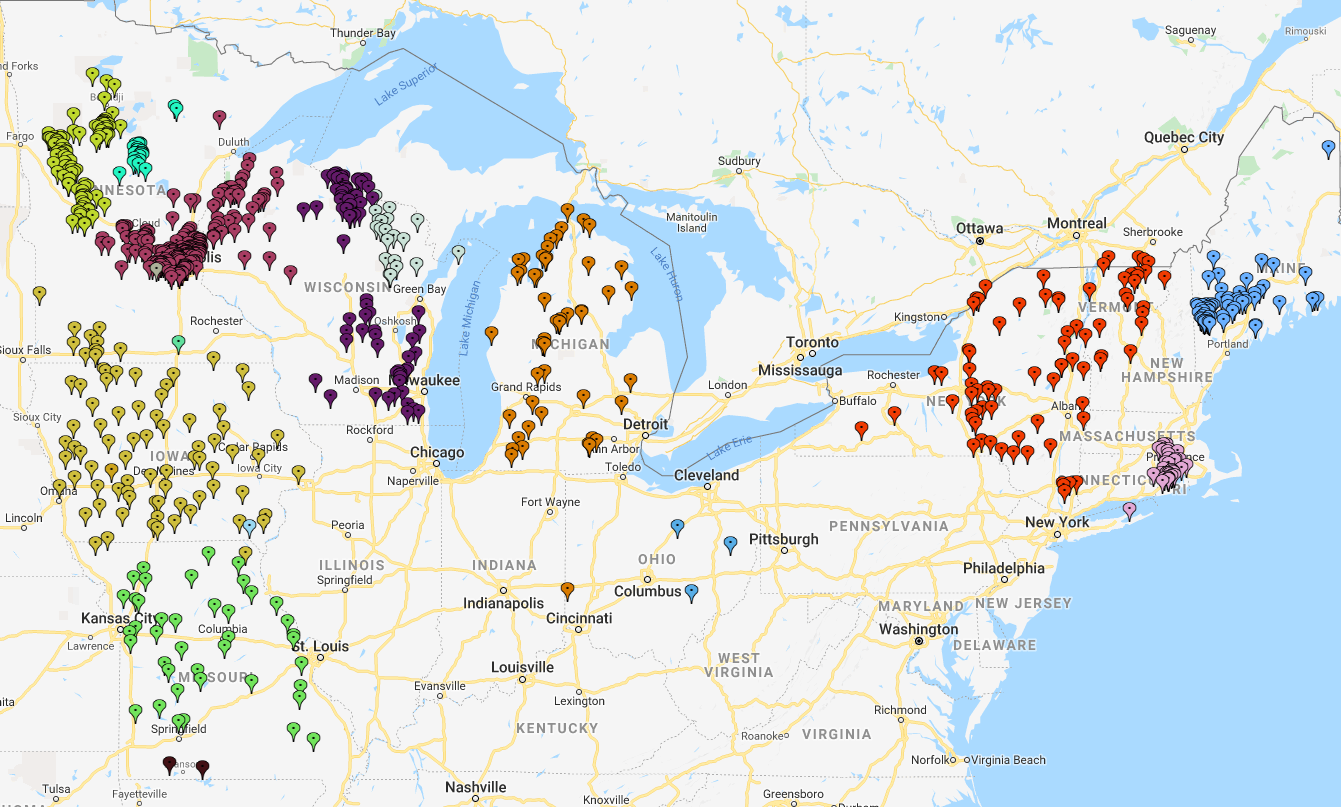}
        \caption{2002}
    \end{subfigure}
    \begin{subfigure}[h]{0.33\textwidth}
    \centering
        \includegraphics[width=\textwidth, clip=true]{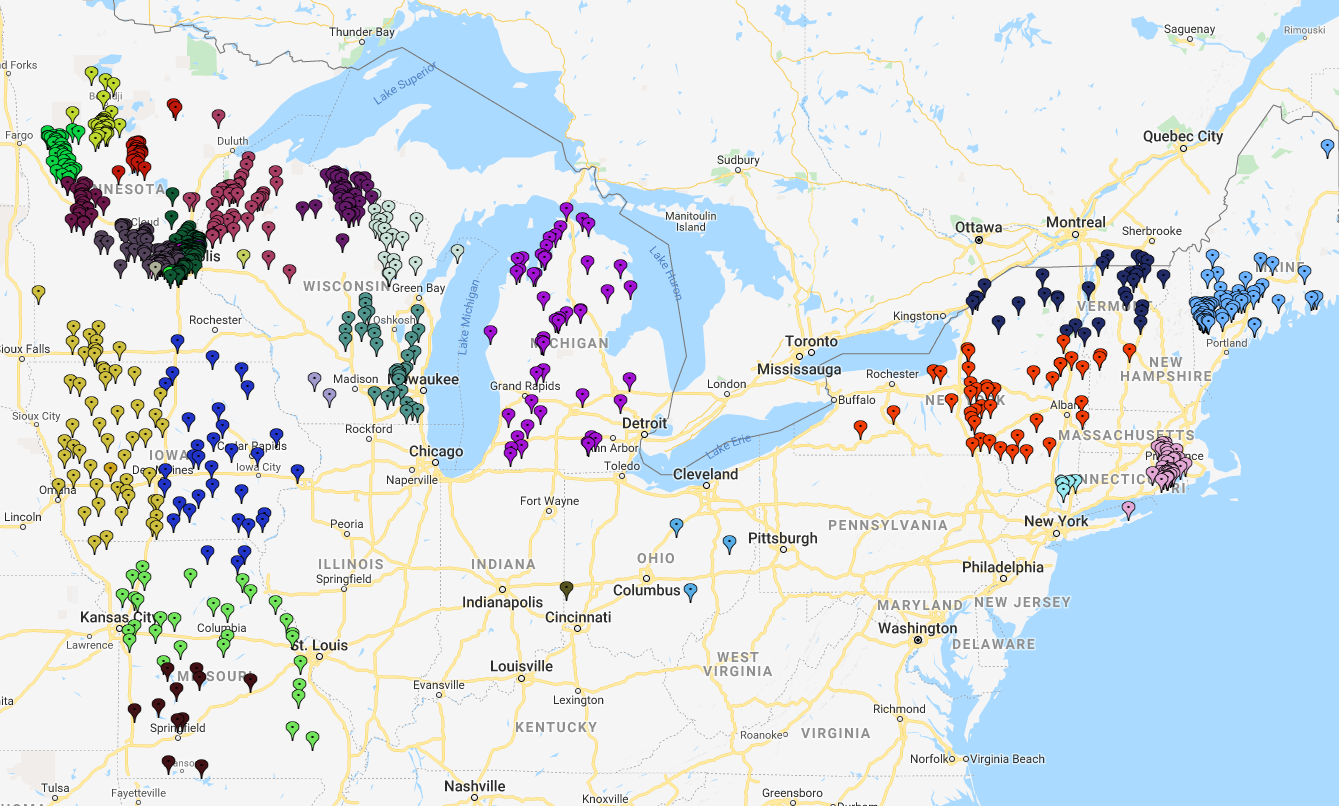}
        \caption{2004}
    \end{subfigure}
    \vspace{-0.15in}
    \caption{\small Evolution of clustering captured by the ST-DANR over years.}
    \vspace{-0.25in}
    \label{fig:lagos_temporal_clustering}
\end{figure*}


\textbf{Test Results.} Once training converges, we predict the rental prices on the test set. To do that, we connect each house in the test set to its 10 nearest neighbors in the training set. We then infer the new model $\bm{x}_j$ by taking the average of the models on its neighbors: $\bm{x}_j = (1/|N(j)|) \sum_{k \in N(j)} \bm{x}_k$. The model $\bm{x}_j$ is further used to estimate the rental price of the corresponding house. Alternatively, one can also infer $\bm{x}_j$ by solving $\min_{\bm{x}_j} \sum_{k \in N(j)} \|\bm{x}_j - \bm{x}_k^*\|_2$, while keeping the models on neighbors fixed. However, we empirically find that simply averaging the neighbors' models performs better for both methods in this particular setting. We compute Mean Squared Error (MSE) over test nodes to evaluate the performance. 
\vspace{-0.1in}
\begin{table}[h!]
\centering
 \begin{tabular}{l c c } 
 \hline
 Method & BAY & SAC \\
 \hline
 Local estimation ($\lambda=0$)     & 0.5984                    & 0.6250 \\ 
 Global estimation ($\lambda > \lambda_{critical}$) & 0.4951    & 0.5403 \\
 rMTFL                              & 0.4774                    & 0.4115 \\
 FORMULA (unweighted)               & 0.4446                    & 0.3503 \\
 FORMULA (weighted)                 & 0.4181                    & 0.3379 \\
 Network Lasso (unweighted)         & 0.4392                    & 0.3273 \\
 Network Lasso (weighted)           & 0.4173                    & 0.3022  \\
 DANR (unweighted)                  & \textbf{0.4106}           & \textbf{0.2978}  \\
 \hline 
 \end{tabular}
 \caption{\small MSE for housing rental price prediction on test set.\vspace{-3ex}}
\label{housing_mse_results}
\end{table}

Table~\ref{housing_mse_results} displays the test results of eight different settings on both datasets. As shown, local \& global estimations and rMTFL method produce high errors for both datasets. We further apply FORMULA and Network Lasso (NL) methods to both weighted and unweighted versions of the networks, while the proposed method is only applied to the unweighted version. Notice that the network weights are irrelevant for the local \& global estimation settings and the rMTFL method. 
Intuitively, both FORMULA and NL performs better on the weighted setting compared to the unweighted setting for both datasets. As shown, the rental price estimation errors achieved by the NL are 0.4173 (weighted) vs. 0.4392 (unweighted) for BAY and 0.3022 (weighted) vs. 0.3273 (unweighted) for SAC. These results suggest that the pre-defined weights on these networks help to learn more accurate models on houses. 

However, we further argue that although such pre-defined weights improve the overall clustering performance, they don't account for more complex clustering scenarios, e.g., two nearby houses falling into different school districts or some houses having higher values compared to their neighbors due to geography, e.g., having a view of the city. That being said, DANR outperforms all the other baselines and achieves the smallest errors for both datasets; 0.4106 for BAY and 0.2978 for SAC. Especially, the DANR (unweighted) even outperforms the weighted version of the baseline approaches by a notable margin. This reveals that the DANR indeed accounts for such heterogeneities in data and provides enhanced clustering of the network. 


Figure~\ref{fig:sac_zoom1} shows examples of two complex scenarios that are captured by the DANR from the SAC network. We use a color code to represent the clusters in the network, where the same colored houses ($i$, $j$) are in consensus on their models ($\bm{x}_i = \bm{x}_j$).
In the left subfigure, the house shown in yellow uses a different model than all of its neighbors. Interestingly, it resides at the border of three different area codes. The area code for this house is 95864, while the area code on its west is 95825 and 95826 on its south-east. Additionally, we observe similar heterogeneous behaviors in some houses that are near creeks, lakes, and rivers. As an example, Figure~\ref{fig:sac_zoom1} (right) displays a creek side house (colored in blue), which again uses a different underlying model from its neighbors.

\vspace{-0.1in}
\subsection{Spatio-Temporal: Water Quality Estimation of Northeastern US Lakes} \label{dataset_lake}

We now evaluate our method in spatio-temporal setting, where the aim is to dynamically estimate the water quality of Northeastern US lakes over years. We follow the same procedure in \S\ref{dataset_housing}, but have an additional temporal regularization term in our objective that allows nodes to obtain signals from past snapshots of the network, along with their neighbors in the current snapshot. 

\textbf{Dataset and Network.} We experiment with the geospatial and temporal dataset of lakes in 17 states in the Northeastern United States, called LAGOS~\cite{soranno2017lagos}. The dataset holds extensive information about the physical characteristics, various nutrient concentration measurements (water quality), ecological context (surrounding land use/cover, climate etc.), and location of lakes; from which we select the following available set of features: \emph{\{max depth, surface area, elevation, annual mean temperature, \% of agricultural land,  \% of urban land,  \% of forest land, \% of wetland\}}. The water quality metric to estimate is the summerly mean \emph{chlorophyll concentration} of the lakes. We represent the feature vector with $\bm{w} \in \mathbb{R}^8$ and the water quality score with $y \in \mathbb{R}$. 

During our experiments, we consider a 10-year period between 2000 and 2010. Due to sparsity in the data, we allow 2 years range between the two consecutive snapshots of the network. This results in total of 1039 lakes with all the aforementioned features and the water quality measurements available for each of the selected years. After randomly splitting 20\% of the lakes for testing, we build our network by using the latitude/longitude information of lakes, where each lake (node) is connected to 10 nearest lakes. 

\textbf{Optimization Problem.} 
After constructing the network, we now aim to \emph{dynamically} estimate the water quality ($y_{i,t}$) of lakes by using their feature vectors ($\bm{w}_{i,t}$). More formally, for each year $t \in [2000, 2002, \cdots, 2010]$, we learn a model $\bm{x}_{i,t} \in \mathbb{R}^8$ per node by solving the following spatio-temporal problem in a {\it streaming} fashion: \hfill(4.16)
\vspace{-3ex}
{\small 
    \begin{align}
        \min_{\bm{x,\alpha,\beta}}~ \sum_{i \in \mathcal{V}} f(\bm{x}_{i,t}, \bm{w}_{i,t}, y_{i,t}) + \lambda_1 \cdot \mathcal{R}_{S}^t(\bm{x}, \bm{\alpha}) + \lambda_2 \cdot \mathcal{R}_{T}^t(\bm{x}, \bm{\beta}) \nonumber
    \end{align}
}
\vspace{-0.18in}

\noindent where $f$ is the linear regression objective and $\mathcal{R}_S$ is the DANR term applied on the spatial edges. $\mathcal{R}_T$ is the temporal regularization term, for which we consider two formulations as described next.
 
\textbf{Test results.}  
For each year, we first solve the spatial problem (Eq.~(4.16) without the temporal term $\mathcal{R}_T$) 
and report the Mean Squared Errors in Table~\ref{lakes_temporal_results}. Note that the test nodes are again inferred by averaging the neighbors' models in the current snapshot. Later, in order to successfully assess the improvements gained by the temporal discrepancy-buffering variables ($\bm{\beta}$), we solve the above spatio-temporal problem with three different temporal regularizers: 

\noindent \textbf{DANR+T-SON: } DANR with temporal sum-of-norms regularizer where $\mathcal{R}_T^t(\bm{x}, \cdot)$ = $\sum_{i \in \mathcal{V}} \|\bm{x}_{i,t} - \hat{\bm{x}}_{i,t-1}\|_2$

\noindent \textbf{DANR+T-SOS: } DANR with temporal sum-of-squares regularizer where $\mathcal{R}_T^t(\bm{x}, \cdot)$ = $\sum_{i \in \mathcal{V}} \|\bm{x}_{i,t} - \hat{\bm{x}}_{i,t-1}\|_2^2$

\noindent \textbf{ST-DANR: } Spatio-temporal discrepancy-aware network regularizer where $\mathcal{R}_T^t(\bm{x}, \bm{\beta})$ = $\mu_2 \cdot \sum_{i \in \mathcal{V}} \|\bm{x}_{i,t} - \hat{\bm{x}}_{i,t-1} + \bm{\beta}_{i,t} \|_2 + (1-\mu_2) \cdot \|\bm{\beta}\|_{1,p}$


DANR+T-SON and DANR+T-SOS approaches simply apply two widely adopted temporal regularizers on the temporal edges (the sum-of-norms regularizer~\cite{lindsten2011clustering} and sum-of-squares reqularizer~\cite{dang2017subnetwork,wang2014low} respectively), while ST-DANR includes the discrepancy-buffering variables on both spatial and temporal edges. Recall that we solve 
the Eq.~(4.16)
in a streaming fashion for simplicity, i.e., each snapshot of the network solves the problem while holding the models learned on the previous snapshot (if available) fixed. Potentially, the performance can further be improved by allowing updates on the previous snapshots.

As we can see from the Table~\ref{lakes_temporal_results}, leveraging the temporal connections between the two consecutive snapshots significantly improves the performance. The DANR+T-SON outperforms the DANR by 8\%-14\%, while DANR+TSOS outperforms the DANR by 10\%-15\%. Moreover, the ST-DANR consistently outperforms both baselines for all the years. This confirms that the proposed formulation accounts for the heterogeneous nature of the temporal networks where some group of nodes exhibits different evolution patterns than the others. 

\begin{table}[h]
\centering
 \begin{tabular}{l | p{.72cm} p{.72cm} p{.72cm} p{.72cm} p{.72cm} p{.72cm} p{.72cm} p{.72cm} p{.72cm} }
 \hline
 Method & 2000 & 2002 & 2004 & 2006 & 2008 & 2010 \\
 \hline
 DANR & 0.8479 & 0.9033 & 0.6384 & 0.6722 & 0.4556 & 0.4113 \\
 \hdashline
 + T-SON & N/A & 0.8308 & 0.5744 & 0.6061 & 0.4109 & 0.3517 \\
 + T-SOS & N/A & 0.8045 & 0.5618 & 0.6045 & 
 0.3978 & 0.3476 \\
 ST-DANR & N/A & 0.8037 & 0.5604 & 0.5866 & 0.3844 & 0.3311 \\
 \hline 
 \end{tabular}
 \caption{MSE for water quality estimation over years.}
 \vspace{-4ex}
\label{lakes_temporal_results}
\end{table}
Figure~\ref{fig:lagos_temporal_clustering} further displays the models (color-coded) learned on three consecutive snapshots, corresponding to years 2000, 2002 and 2004. In 2000, the formed clusters don't go much beyond the boundaries of the states, resulting in coarse clustering of the network. Yet some states such as Missouri and Iowa are grouped into one cluster (shown in green). This suggests that accurately estimating the chlorophyll concentration of lakes based on the selected set of features is indeed challenging. Specifically, the estimation problem depends on several other external factors that potentially affect the volume of plants and algae in lakes; cultural eutrophication, nutrient inputs from human activities to name a few~\cite{soranno2017lagos}. However, as it can be seen from the models learned in 2002 and 2004, the clustering performance improves once we allow temporal regularization to synchronize models between two consecutive snapshots, which is analogous with the reduction in errors over years as reported in Table~\ref{lakes_temporal_results}. Overall, we observe a split of clusters over time, e.g., in Wisconsin, Minnesota, Iowa and New England. Additionally, a dark brown cluster in south Missouri begins to appear in 2002 and further spreads north in 2004. This indicates that by leveraging the temporal connections, the ST-DANR learns improved models and clustering while allowing for heterogeneity in group level evolutions of nodes.

\section{CONCLUSIONS}
We propose discrepancy-aware network regularization (DANR) approach for spatio-temporal networks. By introducing discrepancy-buffering variables, the DANR automatically compensates for inaccurate prior knowledge encoded in edge weights, and enables modeling heterogeneous temporal patterns of network clusters. 
We develop a scalable algorithm based on alternating direction method of multipliers (ADMM) to solve the proposed problem, which enjoys analytic solutions for decomposed subproblems and employs a distributed update scheme on nodes and edges. 
Our experimental results show that DANR successfully captures structural changes over spatio-temporal networks and yields enhanced models by providing robustness towards missing or inaccurate edge weights.

\section{ACKNOWLEDGMENTS} 
This project has received funding from the National Science Foundation under grant IIS-1817046 and the Defense Threat Reduction Agency under award HDTRA1-19-1-0017. 

\bibliographystyle{siam}
\bibliography{ref.bib}

\appendix

\section{RELATED WORKS}

Among existing network-based regularization approaches, there are two major types of edge objectives that are most commonly used: the square-norm objective and the unsquared-norm objective, encouraging different styles of expected solutions. The squared-norm edge objective (i.e., $ \sum_{jk}\omega_{jk}\|\bm{x}_j-\bm{x}_k\|_2^2$), well-known as {\it graph Laplacian regularization} \cite{belkin2006manifold}, assumes underlying models on nodes are smoothly varying as one traverses edges in the network, and accordingly enforces similar but not identical models on linked nodes. Due to the merits of simple computations and good performance, graph Laplacian regularizer has gained popularity in solving problems, such as image deblurring and denoising \cite{pang2017graph}, genomic data analysis \cite{li2008network}, semi-supervised regression \cite{belkin2004regularization},  non-negative matrix factorization \cite{guan2011manifold} and semi-definite programming \cite{weinberger2007graph}. However, the square-norm objective usually induces very dense models. The unsquared-norm objective, known as {\it sum-of-norms} regularizer \cite{hocking2011clusterpath, lindsten2011clustering}, bears some similarity to {\it scale-valued fused lasso} signal approximator \cite{tibshirani2005sparsity} that exists in signal processing applications, for example, {\it total-variation} (TV) regularizer for image denoising, and {\it graph trend filtering} (GTF) regularizer for nonparametric regression. TV regularizer~\cite{rudin1992nonlinear} is developed to penalize both the horizontal and vertical differences between neighboring pixels, which can be thought of as a special scalar-valued network lasso on a 2D grid network. GTF~\cite{wang2015trend} generalizes the successful idea of trend filtering to graphs, by directly penalizing higher order differences across nodes. All regularizers above require correct edge weights so that network structures can be properly used to improve joint estimation. In contrast, our proposed approach allows ambiguous input of network edge weights by introducing the discrepancy-buffering variables. Also, as a variation of SON regularization, our formulation enjoys small model complexity as local consensus is imposed across large-scale networks.

\section{Proof of Lemma 2.1}\label{analytic_solution}

We present the analytic solution for updating $(\bm{u}_{jk}, \bm{u}_{kj})$, introduced in Eq. (2.9). For simplicity, we omit iteration index $l'$, and denote $\bm{a} = \bm{x}_{j}+\bm{\delta}_{jk}^{u}$, $\bm{b} = \bm{x}_{k}+\bm{\delta}_{kj}^{u}$ and $c = \lambda_1 (1-\mu_1)\omega_{jk}$.
\begin{align}
\label{eq-tvnl-tvpp3-admm-adm-u-simple}
\operatorname{min~} \quad g(\bm{u}_{jk},& \bm{u}_{kj}) = \quad c \left \| \bm{u}_{jk} + \bm{\alpha}_{jk} - \bm{u}_{kj} \right \|_2 \nonumber \\ 
& + \frac{\rho_1}{2}\left( \left \| \bm{a}-\bm{u}_{jk}\right \|_2^2 + \left \| \bm{b} - \bm{u}_{kj} \right \|_2^2 \right) 
\end{align}
We discuss the problem in the following two cases:

Case (1): When the objective function in Eq. (\ref{eq-tvnl-tvpp3-admm-adm-u-simple}) is differentiable under the condition of {\small $ \left\|\bm{u}_{jk}+\bm{\alpha}_{jk}-\bm{u}_{kj}\right\|_2 \neq 0$}, taking the sufficient condition of optimality as $\nabla g=0$, we have:
\begin{align}
c \frac{\bm{u}_{jk}+\bm{\alpha}_{jk}-\bm{u}_{kj}}{\left\|\bm{u}_{jk}+\bm{\alpha}_{jk}-\bm{u}_{kj}\right\|_2} - \rho_1(\bm{a} - \bm{u}_{jk}) = 0 \nonumber \\
- c \frac{\bm{u}_{jk}+\bm{\alpha}_{jk}-\bm{u}_{kj}}{\left\|\bm{u}_{jk}+\bm{\alpha}_{jk}-\bm{u}_{kj}\right\|_2} - \rho_1(\bm{b} - \bm{u}_{kj}) = 0 \nonumber
\end{align}
and we can solve above linear system equations to obtain:
\begin{align}
    \bm{u}_{jk} = \frac{(c+\tau \rho_1)\bm{a} + c\bm{b} - c\bm{\alpha}_{jk}}{2c+\tau \rho_1} \nonumber \\
    \bm{u}_{kj} = \frac{c\bm{a} + (c+\tau\rho_1)\bm{b} + c\bm{\alpha}_{jk}}{2c+\tau \rho_1} \nonumber
\end{align}
where we define {\small $\tau$ $\coloneqq$ $ \left\|\bm{u}_{jk}+\bm{\alpha}_{jk}-\bm{u}_{kj}\right\|_2$} which still depends on $\bm{u}_{jk}$ and $\bm{u}_{kj}$. Thus we plug the expressions of $\bm{u}_{jk}$ and $\bm{u}_{kj}$ back into the definition of $\tau$, and achieve the expression of {\small $\tau = \|\bm{a}-\bm{b}+\bm{\alpha}_{jk}\|_2 - 2c/\rho_1$} which is fully comprised of fixed variables. Hence, the solution is:
\begin{align}
    \label{u-update-appendix1}
    \bm{u}_{jk}^{\ast} = (1-\theta)\bm{a} + \theta \bm{b} - \theta \bm{\alpha}_{jk} \nonumber \\
    \bm{u}_{kj}^{\ast} = \theta \bm{a} + (1-\theta)\bm{b} + \theta \bm{\alpha}_{jk}
\end{align}
where {\small $ \theta \coloneqq c/(\rho_1 \|\bm{a}-\bm{b}+\bm{\alpha}_{jk}\|_2) \geq 0$}.\par

Case (2): When the objective function in (\ref{eq-tvnl-tvpp3-admm-adm-u-simple}) is not differentiable, we need to solve the quadratic problem {\small $\operatorname{min} \left \| \bm{a}-\bm{u}_{jk}\right \|_2^2 + \left \| \bm{b} - \bm{u}_{kj} \right \|_2^2 $} with the constraint {\small $\|\bm{u}_{jk}+\bm{\alpha}_{jk}-\bm{u}_{kj}\|_2 = 0 $}. It yields {\small $\bm{u}_{jk}^{\ast} = (\bm{a}+\bm{b}-\bm{\alpha}_{jk})/2$, $\bm{u}_{kj}^{\ast} = (\bm{a}+\bm{b}+\bm{\alpha}_{kj})/2$}, which is in the same form of Eq. (\ref{u-update-appendix1}) if {\small $\theta = 1/2$}. 

To determine which of above differentiable conditions is satisfied, we compare resulting objective values in two cases. Denote $F_1^{\ast}$ and $F_2^{\ast}$ as the optimal objective value in above two cases, we compute their difference as: 
\begin{align}
 \frac{F_1^{\ast} - F_2^{\ast}}{\|\bm{a}-\bm{b}+\bm{\alpha}_{jk}\|_2} = (\theta^2-\frac{1}{4}) \frac{c}{\theta} + \frac{c\tau \rho_1}{2c+\tau\rho_1} \nonumber
\end{align}
From the definition of $\tau$ and $\theta$, we can derive {\small $\tau\rho_1 = c(1/\theta - 2)$}. Combining above, we have:
\begin{align}
 \frac{F_1^{\ast} - F_2^{\ast}}{c\|\bm{a}-\bm{b}+\bm{\alpha}_{jk}\|_2} = -\frac{(\theta - \frac{1}{2})^2}{\theta} \nonumber
\end{align}
which shows that $F_1^{\ast} \leq F_2^{\ast}$ unless {\small $\theta = 1/2$}. Therefore, we can have a unified solution as Eq. (\ref{u-update-appendix1}). 


\section{Proof of Lemma 2.2}

Based on rigorous analysis in \cite{boyd2011distributed} and \cite{eckstein1992douglas}, we know that the convergence to the global optimum is guaranteed for ADMM algorithms on problems of the following form:
\begin{align} \label{admm}
    \operatorname{min}~ h(\bm{x}_0) + g(\bm{z}_0) \qquad ~s.t.~  A\bm{x}_0+B\bm{z}_0=\bm{c}_0
\end{align}
where both $h$ and $g$ need to be convex functions and constraints are linear. Note that our problem in Eq.(2.2) is equivalent to Eq.(2.5). To satisfy conditions of convergence guarantee, we need to fit Eq.(2.5) to the form in Eq.(\ref{admm}).

We let $\bm{x}_0=\bm{x}$ and let $\bm{z}_0=[\bm{u}, \bm{\alpha}]$, and further define $h(\bm{x}_0) = \sum_{j\in \mathcal{V}} f_{j}(\bm{x}_{j})$ and $g(\bm{z}_0)$ $=$ $\lambda_1 (1-\mu_1) \|\bm{\alpha}\|_{1,p}$ $+$ $\lambda_1 \mu_1 \sum_{(j,k)\in \mathcal{E}} \omega_{jk}\|\bm{u}_{jk} $ $+$ $\bm{\alpha}_{jk}-\bm{u}_{kj}\|_{2} $. Next, $h(\bm{x}_0)$ is convex because each cmponent $f_j(\bm{x}_j)$ is required to be convex in our problem setting. $g(\bm{z}_0)$ is also convex in terms of the joint variable $[\bm{u}, \bm{\alpha}]$, by the definition of convexity and the triangle inequality. Lastly, the constraint $ \lbrace [ \bm{u}_{jk}, \bm{u}_{kj} ] $ $=$ $[ \bm{x}_{j}, \bm{x}_{k} ],$ $\forall (j,k) \in \mathcal{E} \rbrace$ is also linear in terms of entries in $\bm{u}$ and $\bm{x}$. Summing up, we can see that Eq.(2.5) fits the form of Eq.(\ref{admm}). Then by theorems in \cite{boyd2011distributed} and \cite{eckstein1992douglas}, Our ADMM approach to solve the problem Eq.(2.2) is guaranteed to converge to the global optimum.


\section{Pseudo-code for Solving DANR Problem} \label{danr_algorithm}

\begin{figure}[h]
    \centering
    \begin{subfigure}[]{0.45\textwidth}
    \centering
        \includegraphics[width=\textwidth, clip=true]{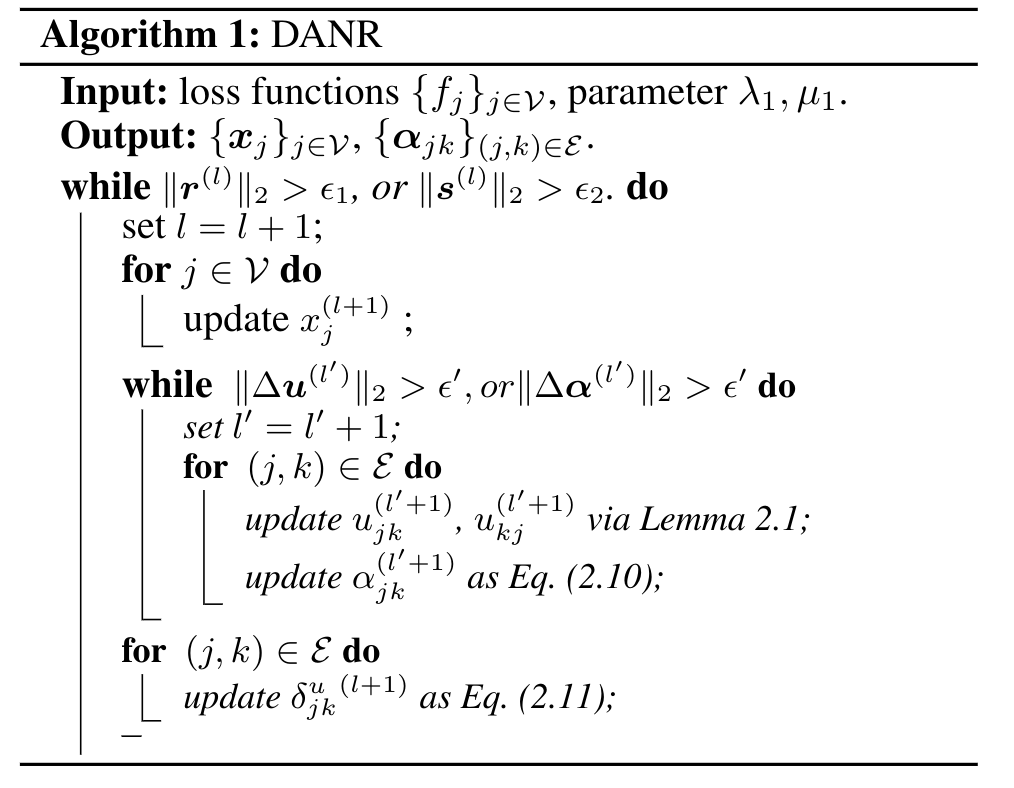}
    \end{subfigure}
\end{figure}


\newpage
\section{Solution of ST-DANR}

\subsection{Batch Formulation} \hfill \\ \par

\noindent Similar to the spatial only case in \S 2, we define $\bm{x}$=$\lbrace \bm{x}_{j,t} \rbrace_{t=1...M}^{j \in V}$ for model parameter vectors on temporal undirected network $\mathcal{G}$, and define consensus variables 
\begin{align}
    \bm{z} &= [\bm{u}, \bm{v}]  \nonumber \\
    &= [\lbrace \bm{u}_{jk,t,0},\bm{u}_{kj,t,0}  \rbrace_{t=1...M}^{(j,k)\in \mathcal{E}_t}, 
    \nonumber \\
    & \hspace{2em} \lbrace \bm{v}_{j,t,1} \rbrace_{t=1...M-1}^{j \in \mathcal{V}_t}, 
    \lbrace \bm{v}_{j,t,2} \rbrace_{t=2...M}^{j \in \mathcal{V}_t}] \nonumber
\end{align}
as copies of $\bm{x}$ in the corresponding spatial penalty term $\mathcal{R}_{S}(\bm{x}, \bm{\alpha})$ and temporal penalty term $\mathcal{R}_{T}(\bm{x}, \bm{\beta})$. Then we rewrite the problem as follows: 
{\small
    \begin{align} \label{eq-tvnl-tvpp3-admm}
       \underset{\bm{x},\bm{z},\bm{\alpha},\bm{\beta}}{\operatorname{min}} &
         \quad \sum_{t=1}^{M} \sum_{j\in \mathcal{V}} f_{j,t}(\bm{x}_{j,t}) \nonumber\\
        & + \lambda_1 \mu_1 \sum_{t=1}^{M} \sum_{(j,k)\in \mathcal{E}} \omega_{jk}^{t}\left\|\bm{u}_{jk,t}+\bm{\alpha}_{jk,t}-\bm{u}_{kj,t}\right\|_{2} \nonumber \\
        & + \lambda_2 \mu_2 \sum_{j\in \mathcal{V}} \sum_{t=1}^{M-1}  \omega_{j}^{t,t+1}\left\|\bm{v}_{j,t,1}+\bm{\beta}_{j,t}-\bm{v}_{j,t+1,2}\right\|_{2}  \nonumber \\
        & + \lambda_1 (1-\mu_1) \|\bm{\alpha}\|_{1,p}+ \lambda_2 (1-\mu_2) \|\bm{\beta}\|_{1,p} \\
        \text{~~s.t.~~} &
        \left[ \bm{u}_{jk,t}, \bm{u}_{kj,t} \right] ~~~= \left[ \bm{x}_{j,t}, \bm{x}_{k,t} \right], \quad t = 1,\ldots,M \nonumber \\ 
        & \left[ \bm{v}_{j,t,1}, \bm{v}_{j,t+1,2} \right] = \left[ \bm{x}_{j,t}, \bm{x}_{j,t+1} \right], \quad t=1,\ldots,M-1 \nonumber
    \end{align}
}\par

\noindent The iterative scheme of ADMM under the above setting is modified as follows, with $l$ denoting the iteration index:\par
{\small
    \begin{align}
        \label{eq-tvnl-tvpp3-admm-scheme}
        \begin{rcases}
            &\bm{x}^{(l+1)}=\underset{\bm{x}}{\operatorname{argmin~}}
            \mathcal{L}_{\rho_1,\rho_2}(\bm{x},(\bm{z},\bm{\alpha},\bm{\beta},\bm{\delta})^{(l)})  \\
            &(\bm{z},\bm{\alpha},\bm{\beta})^{(l+1)}=\underset{\bm{z},\bm{\alpha},\bm{\beta}}{\operatorname{argmin~}}
            \mathcal{L}_{\rho_1,\rho_2}(\bm{x}^{(l+1)},\bm{z},\bm{\alpha},\bm{\beta},\bm{\delta}^{(l)})  \\
            &\bm{\delta}^{(l+1)} = \bm{\delta}^{(l)} + (\tilde{\bm{x}}^{(l+1)}-\bm{z}^{(l+1)})
        \end{rcases}
    \end{align}
}

\noindent where {\small $\tilde{\bm{x}}$=$[\lbrace \bm{x}_{j,t},\bm{x}_{k,t} \rbrace_{t=1...M}^{(j,k)\in \mathcal{E}_t}$,$\lbrace \bm{x}_{j,t} \rbrace_{t=1...M-1}^{j \in \mathcal{V}_t}$, $\lbrace \bm{x}_{j,t} \rbrace_{t=2...M}^{j \in \mathcal{V}_t}]$} is composed of replicated elements in $\bm{x}$. \par

\textbf{$\bm{x}$-Update}.  Now we need to decompose the problem of minimizing $\bm{x}$ into separately minimizing $\bm{x}_{jt}$ for each node $j_t$ :
    \begin{align} 
    \label{eq-tvnl-tvpp3-admm-x-detail}
        & \bm{x}_{j,t}^{(l+1)} =\underset{\bm{x}_{j,t}}{\operatorname{argmin~}} 
        \Big( f_{j,t}(\bm{x}_{j,t})+\frac{\rho_1}{2}\sum_{k \in N_{j,t}}  \|\bm{x}_{j,t} - \bm{u}_{jk,t}^{(l)} + \bm{\delta}_{jk,t}^{u(l)} \|_{2}^{2}    \nonumber \\
        + & \frac{\rho_2}{2}  \|\bm{x}_{j,t}-\bm{v}_{j,t,1}^{(l)} + \bm{\delta}^{v(l)}_{j,t,1} \|_{2}^{2}
        + \frac{\rho_2}{2} \|\bm{x}_{j,t}-\bm{v}_{j,t,2}^{(l)} + \bm{\delta}^{v(l)}_{j,t,2} \|_{2}^{2} \Big)  
    \end{align}

\textbf{($\bm{z},\bm{\alpha},\bm{\beta}$)-Update}. The ($\bm{z},\bm{\alpha},\bm{\beta}$)-update step in Eq.~(\ref{eq-tvnl-tvpp3-admm-scheme}) can be decoupled into separate updates for a spatial ($\bm{u},\bm{\alpha}$)-update and a temporal ($\bm{v}, \bm{\beta}$)-update. For spatial ($\bm{u},\bm{\alpha}$)-update, we have:
{\small 
\begin{align}
        &(\bm{u}_{jk,t}^{(l'+1)}, \bm{u}_{kj,t}^{(l'+1)}) =
        {\operatorname{argmin}} \Big( \lambda_1 \mu_1 \omega_{jk}^{t}\|\bm{u}_{jk,t}+\bm{\alpha}_{jk,t}^{(l')}-\bm{u}_{kj,t}\|_{2} \nonumber \\
          & \hspace{8em}+  \frac{\rho_1}{2} \|\bm{x}_{j,t}^{(l+1)}-\bm{u}_{jk,t} + \bm{\delta}^{u(l)}_{jk,t}\|_{2}^{2}  \nonumber \\
          & \hspace{8em}+  \frac{\rho_1}{2}\|\bm{x}_{k,t}^{(l+1)}-\bm{u}_{kj,t} + \bm{\delta}^{u(l)}_{kj,t} \|_{2}^{2} \Big) \nonumber \\
        &\bm{\alpha}_{jk,t}^{(l'+1)} = {\operatorname{{\footnotesize argmin}}} \Big(\mu_1 \omega_{jk}^{t}\|\bm{u}_{jk,t}^{(l'+1)}+\bm{\alpha}_{jk,t}-\bm{u}_{kj,t}^{(l'+1)}\|_{2} \nonumber \\
        & \hspace{11em} + (1-\mu_1) \| \bm{\alpha}_{jk,t} \|_p  \Big) \nonumber
\end{align}
}
We can follow the same path to solve the temporal ($\bm{v}, \bm{\beta}$)-update, which is omitted here.

\textbf{$\bm{\delta}$-Update}. In the last step of our ADMM updating scheme, we have fully independent update rules for each scaled dual variable {\small $\bm{\delta}_{jk,t}^u$}, {\small $\bm{\delta}_{j,t,1}^v$} and {\small $\bm{\delta}_{j,t,2}^v$} as follows:
{\small
    \begin{align}
        \label{eq-tvnl-tvpp2u-acsx-admm-steps-delta}
        \begin{rcases}
            {\bm{\delta}_{jk,t}^u}^{(l+1)} &= {\bm{\delta}_{jk,t}^u}^{(l)} 
                                    + ( \bm{x}_{j,t}^{(l+1)}-\bm{u}_{jk,t}^{(l+1)}) \qquad\\
            {\bm{\delta}_{j,t,1}^v}^{(l+1)} &= {\bm{\delta}_{j,t,1}^v}^{(l)}
                                    + ( \bm{x}_{j,t}^{(l+1)}-\bm{v}_{j,t,1}^{(l+1)}) \qquad\\
            {\bm{\delta}_{j,t+1,2}^v}^{(l+1)} &= {\bm{\delta}_{j,t+1,2}^v}^{(l)} 
                                    + ( \bm{x}_{j,t+1}^{(l+1)}-\bm{v}_{j,t+1,2}^{(l+1)}) \qquad
        \end{rcases}
    \end{align}
}


\subsection{Streaming Case Formulation} \hfill \\ \par

\noindent Following the same notions in batch formulation of ST-DANR, we formally present the streaming case formulation:

\begin{align} 
  \underset{\bm{x},\bm{\alpha},\bm{\beta}}{\operatorname{min}} &
     \quad \sum_{t=\tau+1}^{\tau+m} \sum_{j\in \mathcal{V}} f_{j,t}(\bm{x}_{j,t}) \nonumber \\
    & + \lambda_1 (1-\mu_1) \|\bm{\alpha}\|_{1,p}+ \lambda_2 (1-\mu_2) \|\bm{\beta}\|_{1,p} \nonumber \\
    & + \lambda_1 \mu_1 \sum_{t=\tau+1}^{\tau+m} \sum_{(j,k)\in \mathcal{E}} \omega_{jk}^{t}\left\|\bm{x}_{jk,t}+\bm{\alpha}_{jk,t}-\bm{x}_{kj,t}\right\|_{2} \nonumber \\
    & + \lambda_2 \mu_2 \sum_{j\in \mathcal{V}} \Big( 
    \omega_{j}^{\tau,\tau+1}\left\|\hat{\bm{x}}_{j,\tau,1}+\bm{\beta}_{j,\tau}-\bm{x}_{j,\tau+1,2}\right\|_{2} \nonumber \\ 
    & \hspace{4em} + \sum_{t=\tau+1}^{\tau+m-1}  \omega_{j}^{t,t+1}\left\|\bm{x}_{j,t,1}+\bm{\beta}_{j,t}-\bm{x}_{j,t+1,2}\right\|_{2} \Big) \nonumber
\end{align}

\noindent where $ \| \bm{\alpha} \|_{1,p} = \sum_{jk \in \mathcal{E}} \sum_{t=\tau+1}^{\tau+m} \| \bm{\alpha}_{jk,t} \|_p $ and $ \| \bm{\beta} \|_{1,p} = \sum_{j \in \mathcal{V}} \sum_{t = \tau}^{\tau+m-1} $ $\| \bm{\beta}_{j,t} \|_p $. We then fit the problem to classic two-block ADMM framework as we did for ST-DANR in batch formulation:
\begin{small}
    \begin{align} \label{eq-tvnl-streaming-admm}
      \underset{\bm{x},\bm{u},\bm{v},\bm{\alpha},\bm{\beta}}{\operatorname{min}}~
        & \sum_{t=\tau+1}^{\tau+m} \sum_{j\in \mathcal{V}} f_{j,t}(\bm{x}_{j,t})  \\
        & + \lambda_1 (1-\mu_1) \|\bm{\alpha}\|_{1,p}+ \lambda_2 (1-\mu_2) \|\bm{\beta}\|_{1,p} \nonumber \\
        & + \lambda_1 \mu_1 \sum_{t=\tau+1}^{\tau+m} \sum_{(j,k)\in \mathcal{E}} \omega_{jk}^{t}\left\|\bm{u}_{jk,t}+\bm{\alpha}_{jk,t}-\bm{u}_{kj,t}\right\|_{2} \nonumber \\
        & + \lambda_2 \mu_2 \sum_{j\in \mathcal{V}} \Big( 
        \omega_{j}^{\tau,\tau+1}\left\|\hat{\bm{x}}_{j,\tau}+\bm{\beta}_{j,\tau}-\bm{v}_{j,\tau+1,2}\right\|_{2} \nonumber \\ 
        & \hspace{2em} + \sum_{t=\tau+1}^{\tau+m-1}  \omega_{j}^{t,t+1}\left\|\bm{v}_{j,t,1}+\bm{\beta}_{j,t}-\bm{v}_{j,t+1,2}\right\|_{2} \Big) \nonumber \\
        \text{~~s.t.~~} &
        \left[ \bm{u}_{jk,t}, \bm{u}_{kj,t} \right] ~~~= \left[ \bm{x}_{j,t}, \bm{x}_{k,t} \right], \quad t = \tau+1,\ldots,\tau+m \nonumber \\
        & \left[ \bm{v}_{j,t,1}, \bm{v}_{j,t+1,2} \right] = \left[ \bm{x}_{j,t}, \bm{x}_{j,t+1} \right],~ t=\tau+1,\ldots,\tau+m-1 \nonumber
    \end{align}
\end{small}where $\hat{\bm{x}}_{\tau}$ is the estimated models at $\tau$-th snapshot. Note that the existence of $\bm{\beta}_{j,\tau}$ and $\bm{v}_{j,\tau+1,2}$ causes a difference between Eq. (\ref{eq-tvnl-streaming-admm}) and the batch case formula in Eq. (\ref{eq-tvnl-tvpp3-admm}), which leads to a modification of updates in $(\bm{v},\bm{\beta})$-updates w.r.t. these two variables:
\begin{small}
    \begin{align}
        & (\bm{v}_{j,\tau+1,2}^{(l+1)}, \bm{\beta}_{j,\tau}^{(l+1)}) =  
        \operatorname{argmin~} \Big(  \lambda_2 \mu_2 \omega_{j}^{\tau,\tau+1}\left\|\hat{\bm{x}}_{j,\tau}+\bm{\beta}_{j,\tau}-\bm{v}_{j,\tau+1,2}\right\|_{2} \nonumber \\
        & + \frac{\rho_2}{2}\|\bm{x}_{j,\tau+1}^{(l+1)}-\bm{v}_{j,\tau+1,2} + \bm{\delta}^{v(l)}_{j,\tau+1,2} \|_{2}^{2}
        + \lambda_2 (1-\mu_2) \| \bm{\beta}_{j,\tau} \|_p \Big)
        \nonumber
    \end{align}
\end{small}This subproblem can still be solved by alternative minimization with following iterative scheme:
\begin{small}
    \begin{align}
        & \bm{v}_{j,\tau+1,2}^{(l'+1)}  = \operatorname{argmin~} \Big(  \lambda_2 \mu_2 \omega_{j}^{\tau,\tau+1}\left\|\hat{\bm{x}}_{j,\tau}+\bm{\beta}_{j,\tau}^{(l')}-\bm{v}_{j,\tau+1,2}\right\|_{2} \nonumber \\
        & \hspace{8em} + \frac{\rho_2}{2}\|\bm{x}_{j,\tau+1}^{(l+1)}-\bm{v}_{j,\tau+1,2} + \bm{\delta}^{v(l)}_{j,\tau+1,2} \|_{2}^{2} \Big)
        \nonumber \\
        \label{eq-tvnl-streaming-admm-adm-beta}
        & \bm{\beta}_{j,\tau}^{(l'+1)} = {\operatorname{{\footnotesize argmin}}} \Big(\mu_2\omega_{j}^{\tau,\tau+1}\left\|\hat{\bm{x}}_{j,\tau}+\bm{\beta}_{j,\tau}-\bm{v}_{j,\tau+1,2}^{(l'+1)}\right\|_{2} \nonumber \\
        &  \hspace{8em} + (1-\mu_2) \| \bm{\beta}_{j,\tau} \|_p  \Big) \nonumber
    \end{align}
\end{small}Each step above is a simple optimization problem. Besides, updating steps for all remaining variables remain the same as in solution for batch formulation.

\section{Running Time and Scalability}

\begin{table}[h!]
\centering
 \begin{tabular}{l| r | r | r | r} 
\hline 
\# nodes     &  DANR	& NL &	rMTFL & FORMULA \\ \hline
100	            &  4.18 s	    &  1.41 s	&   3.32 s	&  7.38 s \\
500	            &  15.10 s	&  5.75 s &  	7.88 s	&    24.62 s \\
1000	        &  48.85 s	&  12.36 s &  	53.91 s	 &   62.68 s\\
5000	        &  146.78 s	&  37.99 s &  	 221.52 s	&  446.21 s \\
10000	        &  484.68 s	&  123.81 s &  	-	& - \\
 \hline 
 \end{tabular}
 \caption{\footnotesize Running time on synthetic datasets when we vary the number of nodes in the network. `-' means the method cannot finish.}
\label{housing_mse_results}
\end{table}
To test the scalability of our method, we create synthetic dataset in the same way as in \S 4.1, but varying the number of nodes in the graph from $100$ to $10000$. Moreover, we tune the probability of adding edges within and between clusters so that the degree of each node is maintained as 20 for all graph sizes. We compare the running time of all methods in the table above. The rMTFL and FORMULA cannot give a result when the graph has 10000 nodes. DANR and NL are both distributed methods, and therefore their running times increase almost linearly with the increasing number of nodes. Comparing with NL, DANR spends a longer time to get more robust results, which suggests a trade-off between complexity and robustness.

\end{document}